\documentclass[runningheads]{llncs}

\usepackage{eccv}

\usepackage{eccvabbrv}

\usepackage{graphicx}
\usepackage{booktabs}

\usepackage{bm}
\usepackage{graphicx}
\usepackage{url}
\usepackage[dvipsnames]{xcolor}
\usepackage{xspace}
\usepackage{amssymb}
\usepackage{amsmath}
\usepackage[colorlinks=true]{hyperref} 
\usepackage[capitalize]{cleveref} 
\usepackage{algorithm}
\usepackage{algorithmicx}
\usepackage{algpseudocode}
\usepackage{wrapfig}
\usepackage{booktabs}
\usepackage{multirow}
\usepackage[table]{xcolor}
\usepackage{pifont}
\usepackage{orcidlink}
\usepackage{makecell}
\usepackage{caption}
\usepackage{color}

\definecolor{catgray}{gray}{0.9}  
\definecolor{linkc}{rgb}{0, 0.44, 0.74}
\definecolor{eqc}{rgb}{1, 0, 0}
\definecolor{newcitecolor}{rgb}{0,0.6,0}
\usepackage{graphicx}
\usepackage[accsupp]{axessibility}  
\usepackage{enumitem}
\usepackage{listings}
\usepackage{tcolorbox}
\tcbuselibrary{listings}
\tcbuselibrary{breakable}

\newcommand{\KVSet}{\mathbf{KV}}
\newcommand{\LNSet}{\mathbf{LN}}
\newcommand{\ModelOuput}{\mathbf{X}_{\theta}}
\newcommand{\NoiseInput}{\mathbf{X}_{T}}
\newcommand{\KVOutput}{\mathbf{kv}}

\newcommand{\vQ}{\mathbf{Q}}
\newcommand{\vK}{\mathbf{K}}
\newcommand{\vV}{\mathbf{V}}

\newcommand{\vS}{\mathbf{S}}

\newcommand{\vP}{\mathbf{P}}

\newcommand{\vO}{\mathbf{O}}

\usepackage{graphicx}
\usepackage{amsmath, amssymb}
\usepackage{algorithm}
\usepackage{algpseudocode}
\usepackage{multirow}
\usepackage[table]{xcolor}
\usepackage{tablefootnote}
\usepackage{subcaption}

\usepackage{wrapfig}
\usepackage{lipsum} 
\usepackage{booktabs} 
\usepackage{adjustbox}

\usepackage[accsupp]{axessibility}  
\usepackage{enumitem}

\usepackage{hyperref}

\usepackage{url}
\newcommand{\annotate}[1]{\textcolor{gray}{{#1}}\xspace}

\usepackage{hyperref}

\usepackage{url}

\newcommand{\blueContent}[1]{{\color{blue}#1}}

\begin{document}

\title{Long-Horizon Streaming Video Generation via Hybrid Attention with Decoupled Distillation} 

\titlerunning{Fast Stream Video Generation with Hybrid Attention}

\author{Ruibin Li\inst{1,2} \and
Tao Yang\inst{2} \and
Fangzhou Ai\inst{2} \and
Tianhe Wu\inst{3} \\
Shilei Wen\inst{2} \and
Bingyue Peng\inst{2} \and
Lei Zhang\inst{1}
}
\authorrunning{R. Li et al.}

\institute{The Hong Kong Polytechnic University \and ByteDance \and City University of Hong Kong}

\maketitle

\begin{abstract}

    Streaming video generation (SVG) distills a pretrained bidirectional video diffusion model into an autoregressive model equipped with sliding window attention (SWA). However, SWA inevitably loses distant history during long video generation, and its computational overhead remains a critical challenge to real-time deployment. In this work, we propose \textbf{Hybrid Forcing}, which jointly optimizes temporal information retention and computational efficiency through a hybrid attention design. First, we introduce lightweight linear temporal attention to preserve long-range dependencies beyond the sliding window. 
    In particular, we maintain a compact key–value state to incrementally absorb evicted tokens, retaining temporal context with negligible memory and computational overhead.
    Second, we incorporate block-sparse attention into the local sliding window to reduce redundant computation within short-range modeling, reallocating computational capacity toward more critical dependencies. 
    Finally, we introduce a decoupled distillation strategy tailored to the hybrid attention design. A few-step initial distillation is performed under dense attention, then the distillation of our proposed linear temporal and block-sparse attention is activated for streaming modeling, ensuring stable optimization.
    Extensive experiments on both short- and long-form video generation benchmarks demonstrate that Hybrid Forcing consistently achieves state-of-the-art performance. Notably, our model achieves  \textbf{real-time}, \textbf{unbounded} $832\times480$ video generation at \textbf{29.5 FPS} on a single NVIDIA H100 GPU \textbf{without quantization or model compression}. The source code and trained models are available at \href{https://github.com/leeruibin/hybrid-forcing}{https://github.com/leeruibin/hybrid-forcing}.

  \keywords{Streaming Video Generation \and Hybrid Attention \and Sparse Attention \and Linear Attention}
\end{abstract}

\section{Introduction}
\label{sec:intro}

\begin{figure}[tb]
  \centering
  \includegraphics[width=\linewidth]{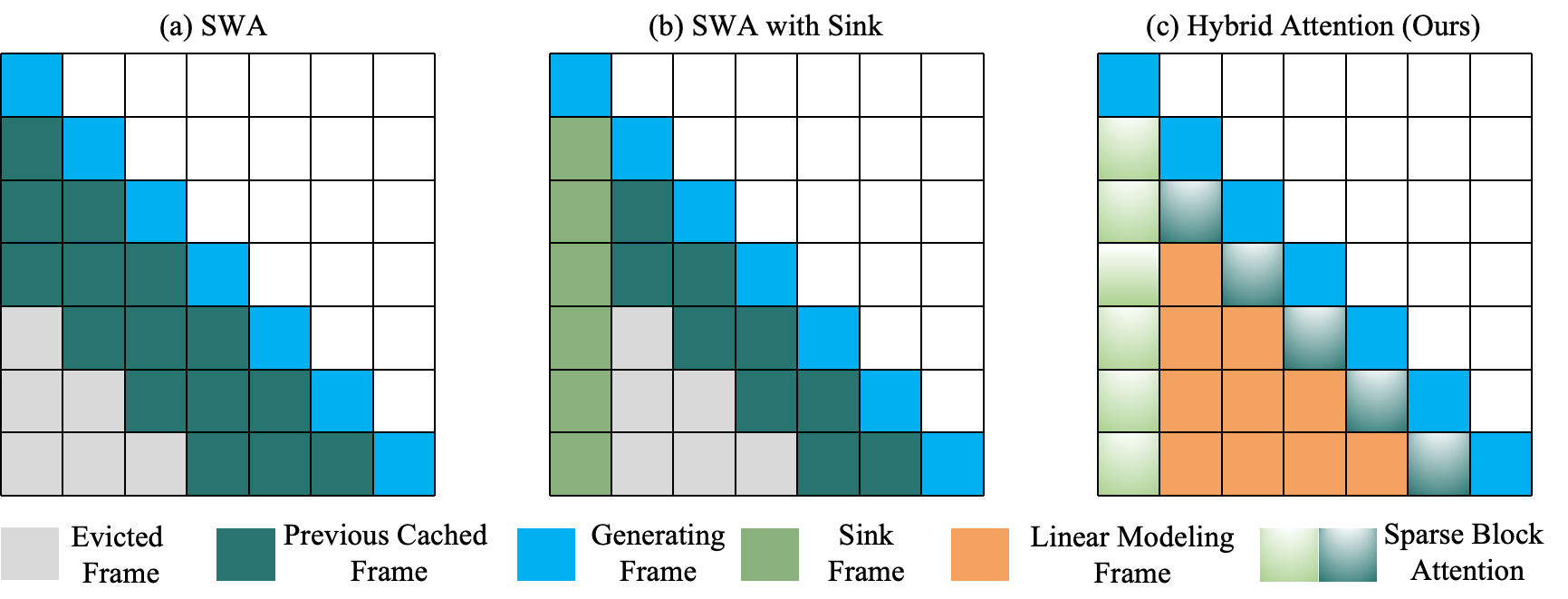 }
  \vspace{-2em}
  \caption{Illustration of our hybrid attention paradigm for SVG. (a) The standard SWA approach only caches the most recent frames, leading to significant error accumulation in long-form generation. (b) While sinking the first frame can reduce early drift, it sacrifices motion diversity and fails to capture long-horizon dependencies effectively. (c) Our hybrid attention combines long-term history retention with efficient local modeling, leveraging low-cost history linear modeling and sparse SWA to enhance performance.}
  \label{fig:attn_scheme}
  \vspace{-2em}
\end{figure}

The rapid scaling of video diffusion models (VDMs) has substantially improved video generation quality, enabling high-fidelity synthesis from text, image, or video inputs~\cite{openai2024sora,google2025veo3,gao2025seedance,zhang2025Waver,opensora2,kong2024hunyuanvideo} to multimodal outputs~\cite{wan21,li2025many,ltx2,gao2025seedance,kling,runway,bruce2024genie}. However, most large-scale VDMs rely on bidirectional attention, restricting generation to short clips due to quadratic memory and computation costs~\cite{zhang2025framepack,yin2025slow,Magi1}. Applying such models to long-form generation is computationally expensive and slow, making real-time or interactive deployment impractical~\cite{huang2025self,yang2025longlive,chen2024diffusion}. 

To reduce latency, recent works convert pretrained bidirectional VDMs into autoregressive (AR) streaming models to generate videos chunk by chunk~\cite{yin2025slow,huang2025self,yang2025longlive,lu2025reward,ji2025memflow}. This adaptation typically involves two stages: (1) ODE-based initialization, where the AR student approximates the reverse diffusion process under block-causal attention masking~\cite{yin2025slow,huang2025self,dong2024flex}; and (2) Distribution Matching Distillation (DMD)~\cite{yin2024one,wang2023prolificdreamer,wu2026diversity,yin2024one}, which compresses generation into only a few inference steps~\cite{huang2025self,yang2025longlive,li2025rorem}. During both distillation and inference, sliding window attention (SWA)~\cite{liu2021swin,huang2025self,zhang2025fast,kodaira2025streamdit} with key–value (KV) caching is employed to enforce causality, as illustrated in \cref{fig:attn_scheme}(a). While SWA significantly improves efficiency, 
as the window advances, earlier frames are progressively evicted from the KV cache, losing access to distant historical context. Consequently, each frame can only condition on a recency-biased temporal window. Errors produced in earlier frames accumulate over time, leading to degraded temporal coherence and unstable long-form generation~\cite{yang2025longlive,cui2025self,dalal2025one}.

Existing streaming approaches mitigate error accumulation using heuristic memory strategies~\cite{liu2025rolling,yang2025longlive,lu2025reward,ji2025memflow}. A representative example is the attention sink method \cite{yang2025longlive}, as shown in \cref{fig:attn_scheme}(b), which retains the initial frame in the KV cache to anchor generation~\cite{xiao2023efficient,lu2025reward}. While this strategy mitigates early drift, retaining a static first-frame sink restricts temporal expressiveness, as repeated attention to the same initial content can induce repetitive patterns and diminished motion diversity. Moreover, informative intermediate frames are often underutilized, losing valuable temporal cues during generation. More recent variants dynamically update the sink based on token similarity~\cite{ji2025memflow}, yet such relevance-based selection is myopic: tokens useful for the current frame may not be informative for future frames, leading to irreversible loss of long-term context. In addition, although these methods are faster than bidirectional models, they can achieve only around 20 FPS~\cite{liu2025rolling,yang2025longlive,ji2025memflow,yesiltepe2025infinity,chen2024diffusion,chen2025skyreels} on an H100 GPU.

We propose Hybrid Forcing, a streaming framework built on a hybrid attention paradigm that unifies long-term history retention and efficient local modeling, as illustrated in \cref{fig:attn_scheme}(c). Our design is based on three principles: (1) compact and learnable long-horizon compression, (2) negligible memory and computational overhead, and (3) seamless integration into the existing streaming pipeline. Firstly, we introduce a linear temporal attention pathway~\cite{chen2025sana,team2025kimi,qiu2025gated} that maintains a compact KV state to summarize evicted tokens and aggregates historical information into a learnable state matrix, enabling scalable long-range dependency modeling without explicit sequence growth. 
Secondly, we incorporate block-sparse attention into SWA for local modeling. Inter-chunk attention maps exhibit substantial sparsity~\cite{zhang2025vsa,zhang2025sla,xi2025sparse,xia2025training}, indicating that many short-range interactions are redundant. By selectively pruning less informative local connections, sparse SWA reduces computational overhead while preserving essential fine-grained refinement. These two components operate synergistically. Sparse local attention removes redundant short-range correlations, and the linear temporal pathway preserves globally aggregated history, effectively reallocating attention capacity toward informative dependencies across time. 
Finally, we adopt a decoupled distillation strategy to fully unlock the potential of the hybrid attention mechanism. We first conduct a short DMD-only phase under dense attention to stabilize autoregressive dynamics and consolidate semantically meaningful token representations. We then activate the proposed hybrid attention for structured streaming training. This decoupled design separates semantic representation learning from structured historical modeling, reducing the impact of noisy and immature historical states on model optimization.

We apply the hybrid attention paradigm to a pretrained VDM with DMD distillation. However, adapting pretrained VDMs to long-horizon streaming introduces additional challenges. First, pretrained models are trained with limited temporal RoPE indices (\eg, 21 frames)~\cite{su2024roformer,wan21}, and naïvely extrapolating to large indices during streaming results in severe distribution shift and unstable generation~\cite{yesiltepe2025infinity,cui2025self}. Additionally, architectural constraints (\eg, capping the temporal RoPE index to 1024) hinder unbounded streaming.
We therefore enforce a consistent upper bound on temporal RoPE indices during both training and inference, substantially improving robustness. Second, DMD-only distillation suffers from mode collapse in long-horizon training~\cite{wu2026diversity,zheng2025large,lu2024simplifying}, manifested as frozen motion or uncontrollable camera drifting. To mitigate this issue, we introduce an auxiliary regularization loss that supervises the student model using the teacher’s rollout results from identical initial noise, encouraging preservation of temporal diversity and motion dynamics.

The main contributions of this paper are summarized as follows:
\begin{itemize}[topsep=0pt, partopsep=0pt, itemsep=0pt, parsep=0pt]
\item We propose \textbf{Hybrid Forcing}, a unified framework for efficient long-horizon streaming video generation. It employs linear attention for long-term history retention, block-sparse attention for efficient local modeling, and a decoupled distillation strategy to separate semantic representation learning from structured historical modeling. Together, these components enable high visual fidelity and low-latency inference.
\item We address two common challenges in existing streaming distillation pipelines, \ie, temporal position embedding mismatch and long-horizon mode collapse, by introducing a consistent relative temporal RoPE constraint during training and inference, together with a teacher-guided regression objective that stabilizes long-term generation.
\item Extensive experiments demonstrate that Hybrid Forcing achieves state-of-the-art performance on both short- and long-horizon video generation benchmarks while significantly reducing inference latency. In particular, Hybrid Forcing achieves \textbf{29.5 FPS} streaming video generation on a single H100 GPU \textbf{without any quantization or model compression}.

\end{itemize}

\section{Related Work}

\noindent\textbf{Long-Duration Video Generation:} 
Much effort has been devoted to extending pretrained video diffusion models (VDMs) for generating longer videos~\cite{jin2024pyramidal, zhang2025framepack, henschel2025streamingt2v, chen2025skyreels}. 
NUWA-XL~\cite{yin2023nuwa} extends diffusion models to handle long sequences by employing a ``diffusion over diffusion'' framework, which generates global keyframes and fills in intermediate frames. LaVie~\cite{wang2025lavie} introduces a cascaded pipeline incorporating joint finetuning, rotary position encoding, and temporal attention. FreeLong~\cite{lu2024freelong} blends temporal frequency components during inference to improve long-duration video generation. Although these models can generate longer videos, they come with substantial computational costs and suffer from degraded video quality over extended durations.

Recently, more attention has been paid to causal autoregressive (AR)-based models for long video generation~\cite{yin2025slow,huang2025self,liu2025rolling,yang2025longlive,lu2025reward,ji2025memflow,cui2025self,yesiltepe2025infinity,zhu2026causal}. CausVid~\cite{yin2025slow} reformulates bidirectional video diffusion into a causal model and leverages DMD~\cite{yin2024one} to compress multi-step models into a few-step one. Self-Forcing~\cite{huang2025self} addresses the train-test mismatch by conditioning on self-generated frames to enhance temporal stability. Rolling Forcing~\cite{liu2025rolling} denoises consecutive frames with progressively increased noise, maintaining a persistent attention-sink cache for global consistency. LongLive~\cite{yang2025longlive} introduces frame-level autoregression and KV recache tuning, further enhancing interactive generation. Relic~\cite{hong2025relic} use highly compressed latent tokens to model long-horizon memory. MAGI-1~\cite{Magi1} progressively denoises chunks, predicting fixed-length segments of consecutive frames. 

Despite the efficiency of these AR-based models, they fall short in \textbf{capturing long-term context due to the use of sliding window attention (SWA)}. Errors introduced in earlier frames accumulate over time, leading to degraded temporal coherence and unstable long-form generation.

\noindent\textbf{Sparse Attention:} Recent works have explored the sparsity in attention maps for video generation~\cite{zhang2025sla, zhang2025vsa, xi2025sparse, xia2025training}. 
Leveraging this property, Sparse Video Generation~\cite{xi2025sparse} methods define a fixed sparse attention mask to reduce the computational cost of attention. AdaSpa~\cite{zhang2025sla} introduces a blockified pattern to efficiently capture the hierarchical sparsity in DiTs, significantly reducing attention complexity. VSA~\cite{zhang2025vsa} adapts the DeepSeek NSA~\cite{yuan2025native} to video DiTs, finetuning the model with newly introduced VSA modules to achieve better performance compared to training-free methods. SLA~\cite{zhang2025sla} introduces a learnable Sparse-Linear Attention mechanism into pretrained VDMs to accelerate inference speed. While these methods have demonstrated efficiency improvements in bidirectional video generation, their effectiveness in SVG remains unproven. Moreover, many of these approaches \textbf{fail to account for the streaming attention paradigm}, which can lead to kernel errors when applying to SVG models.

\section{Methods}

\begin{figure}[tb]
  \centering
  \includegraphics[width=\linewidth]{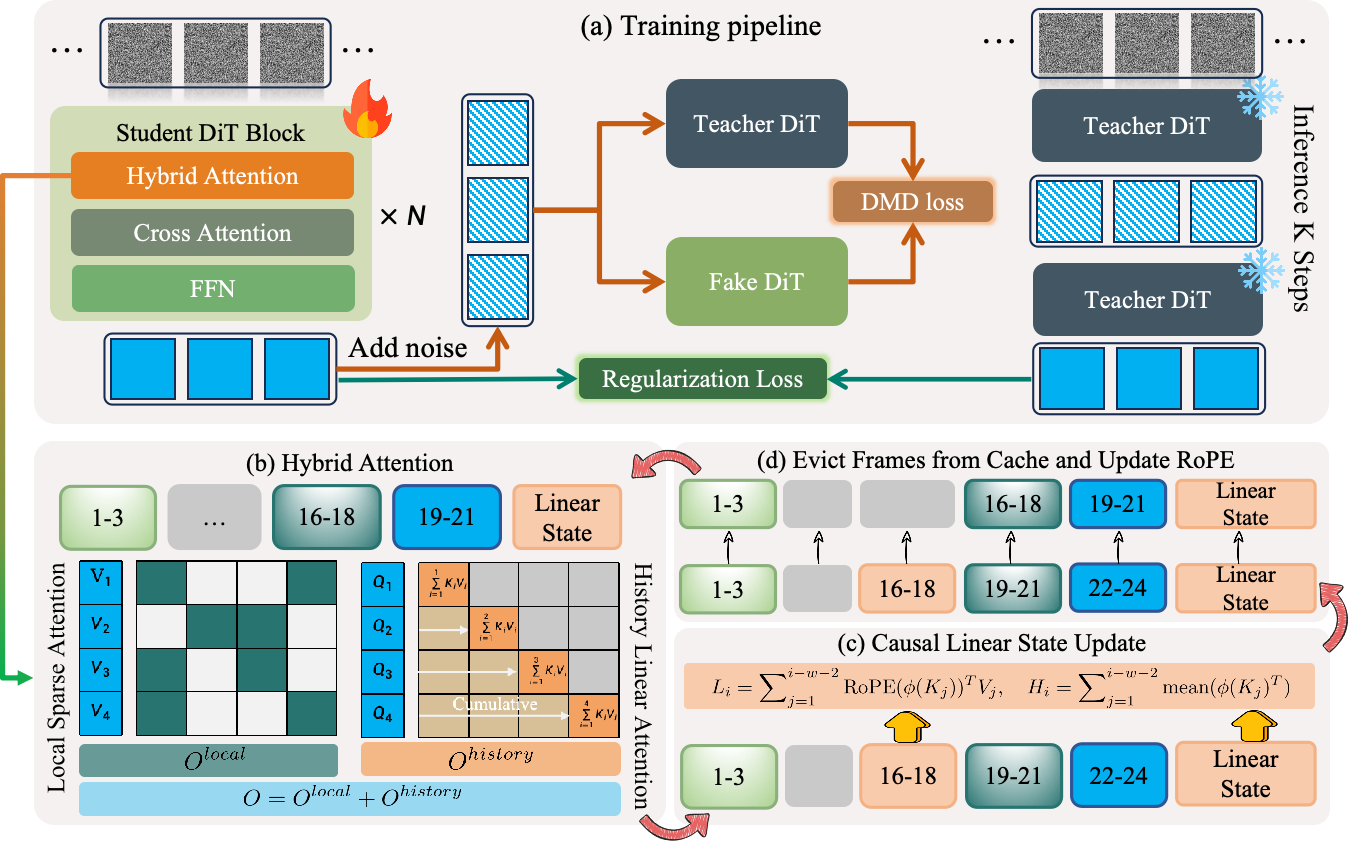 }
  \vspace{-2em}
  \caption{Framework of our training and inference pipeline. (a) An auxiliary regularization loss is introduced to mitigate training mode collapse. (b) The output of our hybrid attention is computed as the sum of sparse local window attention and linear history attention. (c) The linear state is updated prior to evicting outdated frames from the KV cache to preserve long-term information. (d) The temporal RoPE index is constrained within the maximum temporal index of pretrained model.}
  \label{fig:hybrid_framework}
  \vspace{-2em}
\end{figure}

In this section, we first introduce the preliminaries about Streaming Video Generation (SVG), then analyze the limitations of current SVG architectures, including the inability to model long-context history and the inefficiency of  sliding window attention (SWA). We also identify challenges in the training paradigm, such as uncontrollable camera movement and unstable long-horizon video generation. To address these issues, we propose a hybrid attention mechanism and an associated training pipeline, which are illustrated in \cref{fig:hybrid_framework}.

\subsection{Preliminary}

\textbf{Video Diffusion Model (VDM)}: Current VDMs usually adopt the flow-matching paradigm~\cite{lipman2022flow,liu2022flow}, which builds a trajectory from clean video latents $\bm{x}_0 \sim p_{\text{data}}(\bm{x})$  to Gaussian noise $\bm{\epsilon} \sim \mathcal{N}(\bm{0}, \bm{I})$. A point along this trajectory is interpolated as $\bm{x}_t = \alpha_t \bm{x}_0 + \beta_t \bm{\epsilon}$, with $\alpha_t$ and $\beta_t$ controlled by a noise scheduler~\cite{song2020denoising,karras2022elucidating,zhao2023unipc} and $t \in [0,1]$.
The model predicts the velocity field $\bm{v}_{\theta}$ at $\bm{x}_t$. The target velocity is defined as the difference between the noise and the clean data. The loss function is $\mathcal{L}_{FM} = \mathbb{E}_{\bm{x},\bm{\epsilon},t} \left[ \|\bm{v}_{\theta}(\bm{x}_t,t) - (\bm{\epsilon - \bm{x}_0})\|^2 \right]$. New video samples are generated by solving the ODE with the learned velocity field~\cite{lipman2022flow,song2020score}, .

\noindent\textbf{Streaming Video Generation}: Current VDMs are limited to generating short video clips due to the complexity for long video generation~\cite{huang2025self,zhang2025framepack}. SVG generates videos frame-by-frame, factorizing the distribution as $
p(\bm{x}_0^{1:N}) = \prod_{i=1}^{N} p(\bm{x}_0^i \mid \bm{x}_0^{<i})
$, where each $\bm{x}_0^i$ represents a video frame or a chunk of frames. Inference typically uses an SWA mechanism with a fixed window size $w$ to replace the original vanilla attention, limiting the scope to recent frames:
\begin{equation}
    p(\bm{x}_0^{1:N}) = \prod\nolimits_{i=1}^{N} p(\bm{x}_0^i \mid \bm{x}_0^1, \bm{x}_0^{i-w-1:i-1}),
\end{equation}
where $\bm{x}_0^1$ is the sink frame~\cite{yang2025longlive}.
SVGs are often trained through a two-stage distillation pipeline. The first stage uses a pretrained bidirectional diffusion model to generate ODE trajectories, which are then mimicked by a student AR generator $g_{\theta}$ using a causal attention mask. The second stage applies Distribution Matching Distillation (DMD)~\cite{yin2024one} to compress the student model into a few-step generator. The DMD loss is:
\begin{equation}
\nabla_\theta D_\text{KL} = \mathbb{E}_{\epsilon \sim \mathcal{N}(0;\textbf{I})} \big[ - (s_\text{real}(x_t) - s_\text{fake}(x_t)) \frac{\partial g_{\theta}}{\partial \theta} \big],
\label{eq:dmd_kl}
\end{equation}
where $s_{\text{real}}$ and $s_{\text{fake}}$ are score functions~\cite{song2020score}. Despite improvements, two challenges remain: long-range context is lost as the sliding window advances, leading to temporal drift, and the quadratic attention computation within each window remains computationally expensive, hindering real-time deployment.

\subsection{Hybrid Attention}

SVG requires modeling both long-range temporal dependencies and fine-grained local interactions. 
However, standard scaled dot-product attention incurs $\mathcal{O}(N^2 d)$ complexity, making full-history attention infeasible in both memory and computation.
This challenge motivates us to develop a hybrid attention mechanism as illustrated in \cref{fig:hybrid_framework}(b).
Linear attention compresses distant history into a compact representation for modeling global temporal dependencies, while sparse attention preserves fine-grained dependencies within a sliding window for capturing local window interactions. This hybrid approach enables efficient modeling of both long-range and local dependencies, minimizes memory and computational overhead, and can be seamlessly integrated into existing streaming pipelines.

\vspace{+1mm}
\noindent\textit{Linear attention for long-range history modeling.}
Instead of storing all historical frames $\bm{x}_0^{1:i-w-2}$, we maintain a dynamically updated KV state that aggregates evicted frames. The linear state update is performed when a frame is evicted from the cache, as illustrated in \cref{fig:hybrid_framework}(c). The state update rule is:
\begin{equation}
    L_i = \sum\nolimits_{j=1}^{i-w-2} \text{RoPE}(\phi(K_j))^T V_j, \quad H_i = \sum\nolimits_{j=1}^{i-w-2} \text{mean}(\phi(K_j)^T),
\end{equation}
where $\phi(\cdot)$ is the activation function. 
The history linear attention output for frame $i$ is then computed as:
\begin{equation}
    \bm{O}_i^{\text{history}} =
    \text{Linear}\!\left(
    \frac{\text{RoPE}(\phi(Q_i)) L_i}
         {\phi(Q_i) H_i}
    \right).
\end{equation}

Unlike standard linear attention that attends to all frames, we restrict the receptive field to evicted frames only. As the sliding window advances, $L_i$ and $H_i$ are updated online, resulting in constant memory cost independent of video length. This mechanism enables effective long-horizon dependency modeling with negligible overhead. Importantly, because distant information is compressed into a compact state, we can safely reduce the local window size (\eg, from $w=21$ to $w=9$), further lowering inference latency.

\vspace{+1mm}
\noindent\textit{Sparse SWA for efficient local modeling.}
While linear attention captures global consistency, local temporal interactions still require careful modeling. We therefore retain SWA within the recent $w$ frames but accelerate it using block-sparse computation.
Building upon FlashAttention~\cite{dao2022flashattention}, which improves memory efficiency via block-wise tiling, we further exploit the sparsity of attention maps in VDMs~\cite{xi2025sparse,zhang2025sla}. 
Specifically, we compute a block importance score as:
\begin{equation}
    \bm{A}_{ij} = \text{Mean}(\bm{q}_i)\,\text{Mean}(\bm{k}_j)^T,
\end{equation}
where \{$q_i$\}, \{$k_j$\} are blocks from $Q_i$ and $K_{1,i-w:i}$. We retain only the top-$K$ most activated blocks, forming a binary mask $M_{ij}$, whose value is 1 if $\bm{A}_{ij} \in \text{TopK}$ and 0 otherwise. 
Attention is then computed only over active blocks as:
\begin{equation}
    \bm{S}_{ij} = \bm{q}_i \bm{k}_j^T / \sqrt{d}, \quad
    \bm{P}_{ij} = \text{OnlineSoftmax}(\bm{S}_{ij}), \quad
    \bm{O}_i^{\text{local}} = \sum_{j: M_{ij}=1} \bm{P}_{ij} \bm{v}_j.
\end{equation}

\vspace{+1mm}
\noindent\textit{Hybrid attention.}
Finally, the output of our hybrid attention is the sum of compressed global history and sparse local attention:
\begin{equation}
    \bm{O}_i = 
    \bm{O}_i^{\text{history}} +
    \bm{O}_i^{\text{local}}.
\end{equation}

This decomposition naturally separates long-term dependency modeling from short-term interaction modeling, achieving both temporal coherence and real-time efficiency without increasing memory complexity. Specially, the sparse local attention mechanism effectively eliminates redundant short-range correlations, while the linear temporal pathway ensures the preservation of globally aggregated temporal history. This dual approach reallocates attention capacity towards the most informative dependencies over time.

\subsection{Relative RoPE}

Previous work clamps the temporal RoPE index at inference time~\cite{yesiltepe2025infinity}, While this strategy can alleviate extrapolation errors in simple cases and provide unbounded generation capacity, it cannot provide stable long-duration streaming.
To better address these issues, we propose to incorporate \emph{relative RoPE directly into the training process}. 
Specifically, we impose a cap on the maximum temporal RoPE index during both training and inference, and represent temporal positions as relative offsets, as illustrated in \cref{fig:hybrid_framework}(d). The modified temporal RoPE formulation is defined as:
\begin{equation}
    RoPE_{\text{temp}}(i) =
    \begin{cases}
    [i-2,\, i-1,\, i], & \text{if } i \leq \mathcal{T}_0, \\
    [\mathcal{T}_0-2,\, \mathcal{T}_0-1,\, \mathcal{T}_0], & \text{otherwise},
    \end{cases}
\end{equation}
where $\mathcal{T}_0$ denotes the maximum temporal RoPE index used in the pretrained VDM. All spatial coordinates remain unchanged, and the temporal RoPE indices in the KV cache are updated based on the offsets to the current chunk. Meanwhile, we perform long-horizon rollouts during distillation. This exposes the model to realistic streaming settings under relative RoPE, allowing it to adapt to the capped temporal indexing mechanism. As a result, the train--test discrepancy is substantially reduced, leading to improved temporal coherence and stability in extended video generation.

\subsection{Regularization Loss}

While DMD (\cref{eq:dmd_kl}) is effective for accelerating inference, we observe that long-term DMD training in streaming settings often leads to mode collapse~\cite{wu2026diversity,zheng2025large}. In practice, this manifests as reduced motion diversity and systematic artifacts, such as uncontrollable camera drift (e.g., persistent pan-left).
To mitigate this, we introduce a regularization term, which anchors the student generator to the teacher’s denoising trajectory during early timesteps, as shown in \cref{fig:hybrid_framework}(a). Specifically, given a distilled $N$-step student model, we supervise its first generation step using the teacher’s full rollout result from the same initial noise:
\begin{equation}
    \mathcal{L}_{\text{reg}} = 
    \mathbb{E}_{\epsilon}\,
    \big\|
        f\big(g_\theta(\epsilon, 1)\big)
        - \mathrm{rollout}(g_\phi, \epsilon, 1)
    \big\|,
\end{equation}
where $f(\cdot)$ maps the student output at timestep $1$ to the $\bm{x}_0$ space, and $\mathrm{rollout}(\cdot)$ denotes the complete denoising process of the frozen teacher model $g_\phi$.

This formulation is motivated by the observation that the initial noise latent $\epsilon$ lies within the training distribution of both the few-step student and the multi-step teacher. As a result, the teacher’s rollout provides a stable and high-quality reference for the desired generation trajectory, effectively constraining the student against degenerate modes. This supervision is applied entirely in latent space, avoiding the overhead of pixel-space decoding. For the remaining steps, we retain the standard DMD objective to refine visual fidelity. (Note that the rollout noise–video pairs can be precomputed to eliminate rollout latency during training.) Thus, the final loss function is defined as 
$\mathcal{L}_{distill} = \mathcal{L}_{DMD} + \lambda \mathcal{L}_{Reg}$ when $t = 1$, 
and $\mathcal{L}_{distill} = \mathcal{L}_{DMD}$ for all subsequent steps.

\subsection{Decoupled Distillation}
\label{sec:twostage}

We observe that at the early stage of training, tokens projected from ODE-based initialization checkpoints contain limited semantic information. During this phase, dense attention is essential for refining spatial details and suppressing artifacts, as it enables unrestricted information exchange across tokens. Introducing hybrid attention too early at this stage can undesirably restrict information flow and compress noisy, semantically immature states, potentially trapping optimization in suboptimal minima.

Motivated by this observation, we adopt a decoupled distillation strategy that explicitly separates semantic representation learning from structured historical modeling. Instead of jointly learning both objectives from the outset, we first conduct a DMD-only distillation phase under dense attention, without any architectural or attention modifications. This stage focuses purely on forming information-rich token representations, yielding a reliable checkpoint capable of coherent short-horizon generation under reduced sampling steps. After semantic representations are sufficiently structured, we activate the proposed linear temporal attention and block-sparse attention modules. At this point, the model can effectively learn structured long-range dependency modeling and reallocate attention capacity toward informative temporal interactions without suffering from premature distillation. By disentangling representation formation from temporal structuring, the proposed decoupled training paradigm stabilizes optimization and consistently improves performance. A complete description of the algorithm is provided in the \textbf{Supplemental Materials}.

\section{Experiment}

\subsection{Experiment Setting}

\noindent\textbf{Implementation Details.}
Following Self-Forcing~\cite{huang2025self}, we build Hybrid Forcing upon Wan2.1-T2V-1.3B~\cite{wan21}, a pretrained bidirectional text-to-video diffusion model that generates 81-frame videos at a resolution of $832 \times 480$. We initialize our model from the publicly released ODE checkpoint of Self-Forcing to ensure fair comparison with prior streaming baselines~\cite{yang2025longlive,ji2025memflow,liu2025rolling}.

During DMD distillation, text prompts are sampled from a filtered and LLM-augmented version of VidProM~\cite{wang2024vidprom}, with a global batch size of 32. As described in Sec.~\ref{sec:twostage}, training proceeds in two phases: we first conduct 1000 steps under the original Self-Forcing paradigm to establish semantic knowledge, followed by 1000 additional steps with hybrid attention enabled.
All methods operate in a chunk-wise streaming manner, where each chunk contains three latent frames. The sliding window size is set to 9. We retain the first three frames together with a sink frame in the KV cache. Block-sparse attention is implemented using Triton~\cite{triton}, with a top-20\% sparsity ratio. Increasing the sparsity ratio may lead to performance degradation, while yielding only marginal speed improvements in our preliminary experiments. The maximum relative temporal RoPE index is capped at 21 during both training and inference. The regularization loss weight is set to 0.05.
We use AdamW to optimize both the student generator $g_{\theta}$ (learning rate $1.5\times10^{-6}$) and the fake score model (learning rate $2\times10^{-7}$). The generator is updated once for every five fake score updates.

\noindent\textbf{Evaluation Metrics.}
For short-form evaluation, we adopt VBench~\cite{huang2025self}. Following prior work~\cite{huang2025self,yang2025longlive}, we expand the test prompts using Qwen2.5-7B-Instruct~\cite{bai2023qwen} for comprehensive evaluation. For long-form evaluation, we use VBench-Long and generate 30-second videos for quantitative comparison. To ensure fairness, all quantitative comparisons are conducted under identical settings: 30-second duration at 16 FPS. Following Self-Forcing, we additionally report model throughput measured on a single NVIDIA H100 GPU.

\subsection{Results}

\begin{table}[t!]
  \setlength{\tabcolsep}{3.5pt} 
  \caption{
    \textbf{Comparison with state-of-the-arts} open-source video generation models of similar parameter sizes and resolutions. Evaluation scores are calculated on the extended prompt suite of VBench~\cite{huang2025self}. The FPS is reported on a single H100 GPU. The best and 2nd-best results are highlighted in \textbf{bold} and \underline{underline}, respectively. 
  }
  \vspace{-0.5em}
  \label{tab:short}
  \centering
\resizebox{\linewidth}{!}{
\begin{tabular}{lcccccccc}
  \toprule
  \multirow{2}{*}{Model} & \multirow{2}{*}{\#Params} & \multirow{2}{*}{Resolution} & \multirow{2}{*}{\makecell{Throughput\\(FPS) $\uparrow$}} & \multicolumn{5}{c}{Evaluation scores $\uparrow$}\\
  \cmidrule(lr){5-9}
   &  &  &  & Total & Quality & Semantic & Dynamic & Overall Consistency \\
  \midrule
  \rowcolor{catgray}
  \multicolumn{9}{l}{\textit{Diffusion models}}\\
  Wan2.1~\cite{wan21}                         & 1.3B & $832{\times}480$ & 0.78   & \underline{83.37} &\underline{84.30} &79.65 &61&25.50\\
  \midrule
  \rowcolor{catgray}
  \multicolumn{9}{l}{\textit{Multiple-Step Autoregressive Models}}\\
  SkyReels-V2~\cite{chen2025skyreels}            & 1.3B & $960{\times}540$ & 0.49   & 81.97 &83.96 &74.01&37&25.31 \\
  MAGI-1~\cite{Magi1}                      & 4.5B & $832{\times}480$ & 0.19   & 78.88 &81.67 &67.72&42&25.56 \\
  NOVA~\cite{deng2024autoregressive}                  & 0.6B & $768{\times}480$ & 0.88   & 80.31 &80.66 &78.92&46&25.42 \\
  Pyramid~\cite{jin2024pyramidal}          & 2B   & $640{\times}384$ & 6.7    & 80.98   &83.37   &71.42&52&26.08\\
  \midrule
  \rowcolor{catgray}
  \multicolumn{9}{l}{\textit{Few-Step Autoregressive Models}}\\
  CausVid~\cite{yin2025slow}                    & 1.3B & $832{\times}480$ & 15.4   & 82.65   &84.08   &76.91&\textbf{72}&25.72 \\
  Self Forcing~\cite{huang2025self} & 1.3B & $832{\times}480$          & 15.4   & 82.89   &83.90   &78.83&69&\underline{26.57} \\
  Rolling Forcing~\cite{liu2025rolling} & 1.3B & $832{\times}480$ & 15.8    & 82.95   &83.89   &79.22&52&26.46 \\

  Long Live~\cite{yang2025longlive} & 1.3B & $832{\times}480$ & 20.7    &83.05   &83.79   &\underline{80.10}&50&26.29 \\

  MemFlow~\cite{ji2025memflow} & 1.3B & $832{\times}480$ & 18.7    &83.01   &84.22   &78.15&59&26.19 \\
  
  InfRoPE~\cite{ji2025memflow} & 1.3B & $832{\times}480$ & \underline{21.8}    &81.18   &83.47   &74.02&69&26.28 \\
  \midrule
  Hybrid Forcing & 1.3B & $832{\times}480$ & \textbf{29.5}    & \textbf{83.60}   & \textbf{84.45}   &\textbf{80.19}&\textbf{72}&\textbf{26.72} \\
  \bottomrule
\end{tabular}
}
\vspace{-1.5em}
\end{table}

\noindent\textbf{Short-Duration Evaluation.} We compare Hybrid Forcing with state-of-the-art models, including bidirectional diffusion models Wan2.1-1.3B~\cite{wan21}, multi-step autoregressive models SkyReels-V2~\cite{chen2025skyreels}, MAGI-1~\cite{Magi1}, NOVA~\cite{deng2024autoregressive}, Pyramid Flow~\cite{jin2024pyramidal}, and distilled few-step autoregressive models CausVid~\cite{yin2025slow}, Self-Forcing~\cite{huang2025self}, Long Live~\cite{yang2025longlive}, Rolling Forcing~\cite{liu2025rolling}, and MemFlow~\cite{ji2025memflow}.

As shown in Table~\ref{tab:short}, Hybrid Forcing consistently achieves the best performance across all metrics, including Total, Quality, and Semantic scores. Compared with multi-step autoregressive models, our method improves the Dynamic metric by 38.4\% and achieves a $+1.22$ gain in overall semantic and style consistency, even outperforming the strongest multi-step baseline ($+0.58$ for Pyramid Flow). 
Compared with few-step autoregressive approaches, Hybrid Forcing delivers consistently superior overall performance while achieving the highest throughput, improving the throughput by 35\% over the second-best streaming baseline. Notably, methods such as Long Live and MemFlow increase throughput by shrinking the local window size (\eg, from 21 to 12), which significantly compromises long-range consistency. In contrast, even with a smaller window size of 9, Hybrid Forcing preserves global coherence through the linear temporal pathway, demonstrating the effectiveness of hybrid attention. 
Remarkably, Hybrid Forcing surpasses the  bidirectional Wan2.1 model in overall performance while achieving approximately $30\times$ higher throughput under identical hardware settings. Other detailed metrics are provided in the \textbf{Supplemental Materials}. 

\begin{figure}[tb]
  \centering
  \includegraphics[width=\linewidth]{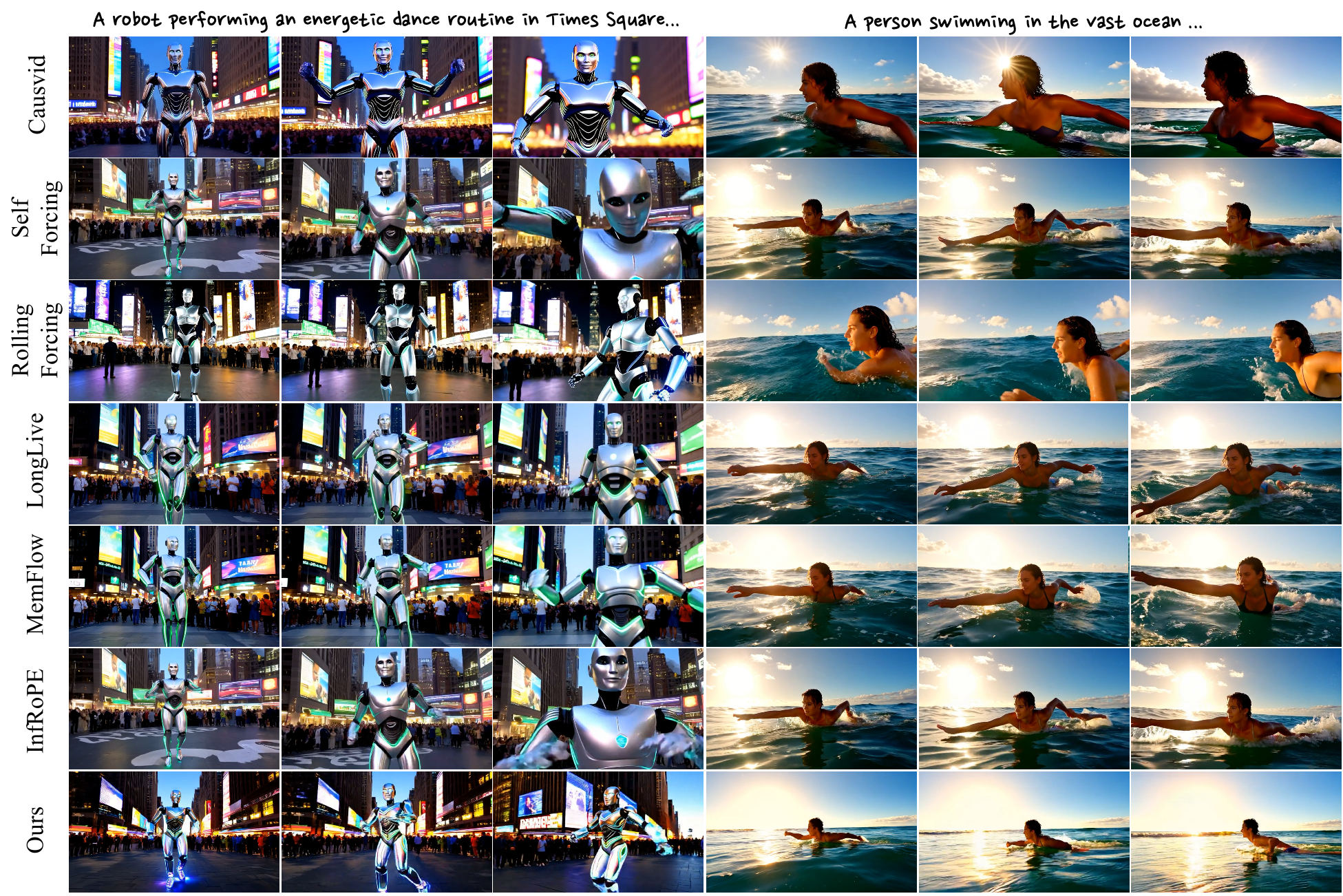  }
  \vspace{-2em}
  \caption{Qualitative results on 5-second videos. Our method exhibits significantly stronger motion dynamics and richer camera movement diversity. Additional comparisons and video demonstrations are provided in the \textbf{Supplemental Materials}.
  }
  \label{fig:visual_result_short}
  \vspace{-1.5em}
\end{figure}

Qualitative comparisons are presented in Fig.~\ref{fig:visual_result_short}. As can be seen, short-window SWA methods LongLive and MemFlow exhibit limited motion dynamics, while long-window SWA baselines CausVid and Self-Forcing capture relatively better motion patterns with higher computation cost and delay. In contrast, our method generates more vivid motion dynamics and diverse camera movements but with a shorter local SWA size, consistent with the quantitative results reported in \cref{tab:short}. More visual results are provided in the \textbf{Supplemental Materials}.

\begin{table}[t]
  \setlength{\tabcolsep}{3.5pt} 
  \caption{
    \textbf{Comparison with state-of-the-art}  open-source few-step video generation models on long-duration inference setting. \textit{Consis.} means Overall consistency. The best and 2nd-best results are highlighted in \textbf{bold} and \underline{underline}, respectively.
  }
  \vspace{-0.5em}
  \label{tab:long}
  \centering
\resizebox{\linewidth}{!}{
\begin{tabular}{lccccccccc}
  \toprule
  \multirow{2}{*}{Model} & \multirow{2}{*}{\#Params} & \multirow{2}{*}{Resolution} & \multirow{2}{*}{\makecell{Throughput\\(FPS) $\uparrow$}} & \multicolumn{5}{c}{Evaluation scores $\uparrow$}\\
  \cmidrule(lr){5-9}
   &  &  &  & Total & Quality & Semantic & Dynamic & Consis. \\
  \midrule
  CausVid~\cite{yin2025slow}                    & 1.3B & $832{\times}480$ & 15.4   &79.67   &83.11   &65.89 &51&22.69  \\
  Self Forcing~\cite{huang2025self} & 1.3B & $832{\times}480$             & 15.4   &81.23   &82.74   &75.20&52&25.23 \\
  Rolling Forcing~\cite{liu2025rolling} & 1.3B & $832{\times}480$         & 15.8   &\underline{82.66}   &\underline{83.81}   &\underline{78.03}&53&\underline{25.77} \\
  Long Live~\cite{yang2025longlive} & 1.3B & $832{\times}480$             & 20.7   &80.12   &82.26   &71.56&34&24.20 \\
  MemFlow~\cite{ji2025memflow} & 1.3B & $832{\times}480$                  & 18.7   &80.70   &83.40   &69.92&61&23.91 \\
  InfRoPE~\cite{yesiltepe2025infinity} & 1.3B & $832{\times}480$          & \underline{21.8}   &82.18   &83.47   &77.02&\underline{62}&25.62 \\
  \midrule
  Hybrid Forcing & 1.3B & $832{\times}480$                             & \textbf{29.5}    &\textbf{83.49}   &\textbf{84.67}   &\textbf{78.78} &\textbf{74}&\textbf{26.02}\\
  \bottomrule
\end{tabular}
}
\vspace{-1.5em}
\end{table}

\vspace{+0.5mm}
\noindent\textbf{Long-Duration Evaluation.} To evaluate long-horizon streaming performance, we extend generation to 126 latent frames.
The evaluation is conducted on VBench-Long~\cite{huang2025self}, and the results are summarized in Table~\ref{tab:long}.
We see that Hybrid Forcing achieves the best overall performance and the highest throughput among all few-step autoregressive baselines. While competing methods typically exhibit noticeable degradation in Quality or Dynamic scores under long-duration generation, our approach maintains stable performance and achieves even improved Quality compared to short-form results. In particular, methods such as CausVid and Self-Forcing show strong dynamics in 5-second generation, but experience significant drops in 30-second settings. Our Hybrid Forcing shows state-of-the-art Dynamic performance and exceeds the second-best method by 19.3\%. Furthermore, our hybrid attention design (short local window + linear history aggregation) consistently outperforms both long-window dense attention models and short-window streaming baselines, demonstrating superior temporal consistency and semantic alignment in extended generation. 

\begin{figure}[tb]
  \centering
  \includegraphics[width=\linewidth]{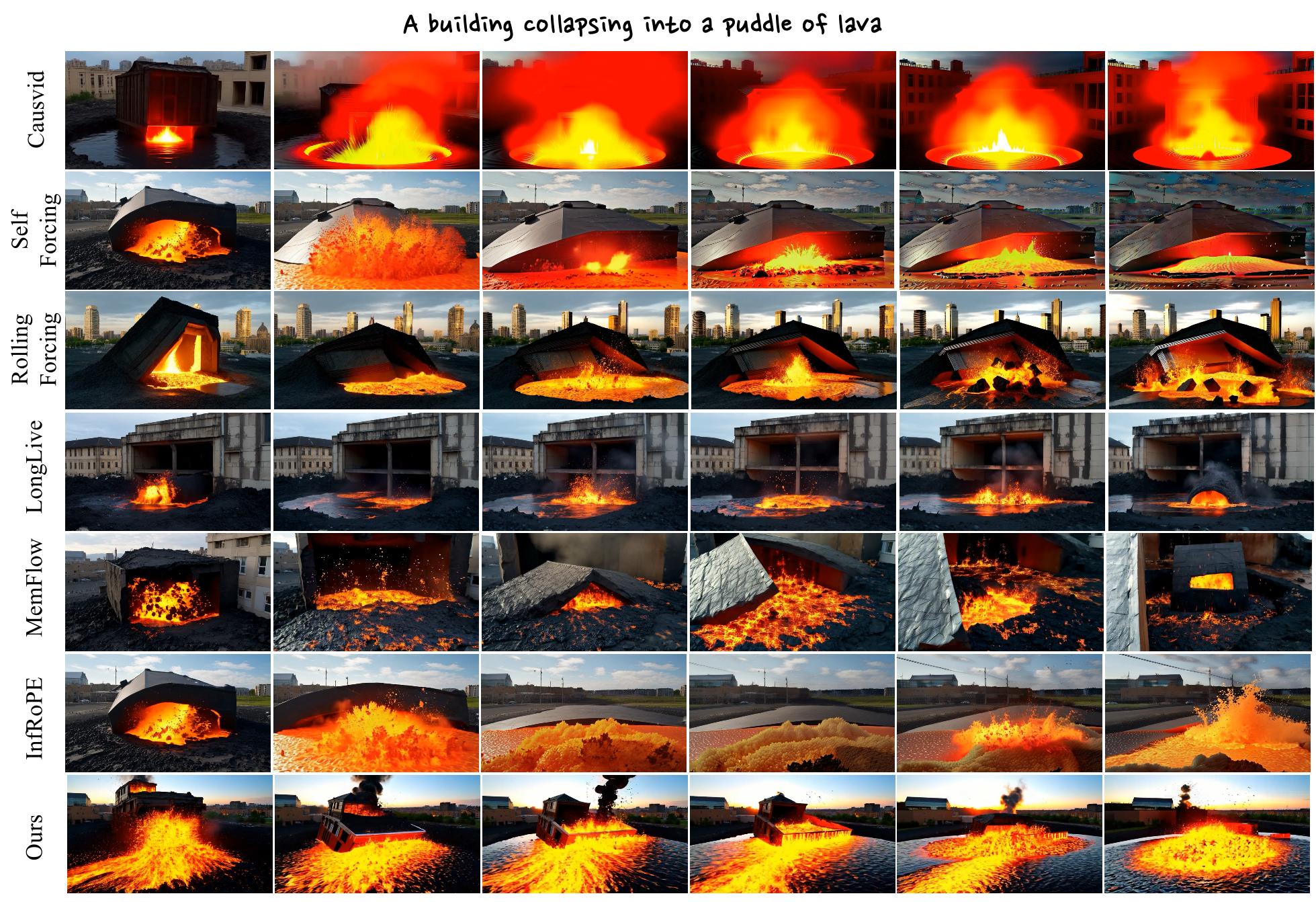 }
  \vspace{-2em}
  \caption{Frame-by-frame qualitative results on 30-second videos. More comparison results and video demos are provided in the \textbf{Supplemental Materials}.}
  \label{fig:visual_result_long}
  \vspace{-0.5em}
\end{figure}

Qualitative long-form results are shown in Fig.~\ref{fig:visual_result_long}. We can see that CausVid exhibits significant error accumulation, leading to progressive saturation drift over time. While Self-Forcing and InfRoPE perform reasonably well in short-duration generation, they experience noticeable degradation in sky regions for longer sequences. LongLive captures minimal scene dynamics, whereas Rolling Forcing and MemFlow show moderate performance. In contrast, our method faithfully reproduces the full “house melt” process with high motion dynamics and visual fidelity. Moreover, our model maintains temporal stability over unbounded horizons, successfully generating videos of more than 10 minutes. Visual evidences  are provided in the \textbf{Supplemental Materials}.

\subsection{Ablation Study}

We conduct ablation studies to disentangle the structural and optimization contributions of Hybrid Forcing. In particular, we examine how the proposed hybrid attention interacts with the decoupled distillation strategy, relative temporal RoPE, and auxiliary regularization. Results are summarized in Table~\ref{tab:ablation}. All variants are distilled for 2000 steps on the same dataset, with a fixed local window size of 9 and 3 sink frames retained in the KV cache .

\noindent\textbf{Hybrid Attention and Decoupled Distillation.}
Introducing hybrid attention alone improves both efficiency and generation quality compared with native short window baseline, increasing throughput from 21.8 to 29.5 and the total score from 80.25 to 81.56. The most pronounced gain appears in the Dynamic metric (53 $\rightarrow$ 67), indicating substantially improved long-horizon motion modeling and reduced error accumulation. This verifies that combining sparse local attention with linear long-range aggregation enhances temporal dependency modeling while maintaining computational scalability.

\begin{table}[t]
  \setlength{\tabcolsep}{3.5pt} 
  \caption{
    Ablation studies on our design. The results are evaluated in the VBench-long suite with 30-seconds inference setting.
  }
  \vspace{-0.5em}
  \label{tab:ablation}
  \centering
\resizebox{\linewidth}{!}{
\begin{tabular}{cccccccccc}
  \toprule
  \multirow{2}{*}{\makecell{Hybrid\\Attention}} & \multirow{2}{*}{\makecell{Relative\\RoPE}} & \multirow{2}{*}{\makecell{Reg\\Loss}} & \multirow{2}{*}{\makecell{Decoupled\\Distillation}} & \multirow{2}{*}{\makecell{Throughput\\(FPS) $\uparrow$}} & \multicolumn{5}{c}{Evaluation scores $\uparrow$}\\
  \cmidrule(lr){6-10}
   &  &  &  &  & Total & Quality & Semantic & Dynamic & Consis. \\
  \midrule
  \ding{55} & \ding{55} & \ding{55} & \ding{55} &21.8 &80.25   &81.65   &74.64 &53&25.70  \\
  \midrule
  \ding{51} & \ding{55} & \ding{55} & \ding{55} &29.5 &81.56   &82.95   &76.01 &67&25.62 \\
  
  \ding{51} & \ding{51} & \ding{55} & \ding{55} &29.5  &81.85   &83.11   &76.82 &64&25.97 \\
  
  \ding{51} & \ding{51} & \ding{51} & \ding{55} &29.5  &82.38   &83.87   &76.41 &65&25.68 \\
  \midrule
  \ding{51} & \ding{51} & \ding{51} & \ding{51} &29.5  &\textbf{83.49}   &\textbf{84.67}   &\textbf{78.78} &\textbf{74}&\textbf{26.02} \\
  \bottomrule
\end{tabular}
}
\vspace{-1.5em}
\end{table}

Specially, the architectural benefit can be fully unlocked when paired with the proposed decoupled distillation strategy. Incorporating decoupled training further boosts the total score to 83.49, with consistent improvements across all metrics, particularly Dynamic (74) and Semantic (78.78). By explicitly separating semantic representation learning from structured historical modeling, the first DMD-only phase consolidates forms information-rich token representations under dense attention. Then subsequent distillation phase can focus on learning structured long-range history compression and sparsification over already coherent tokens. This separation of objectives prevents premature compression of noisy historical states and leads to more stable optimization.

\begin{figure}[tb]
  \centering
  \includegraphics[width=\linewidth]{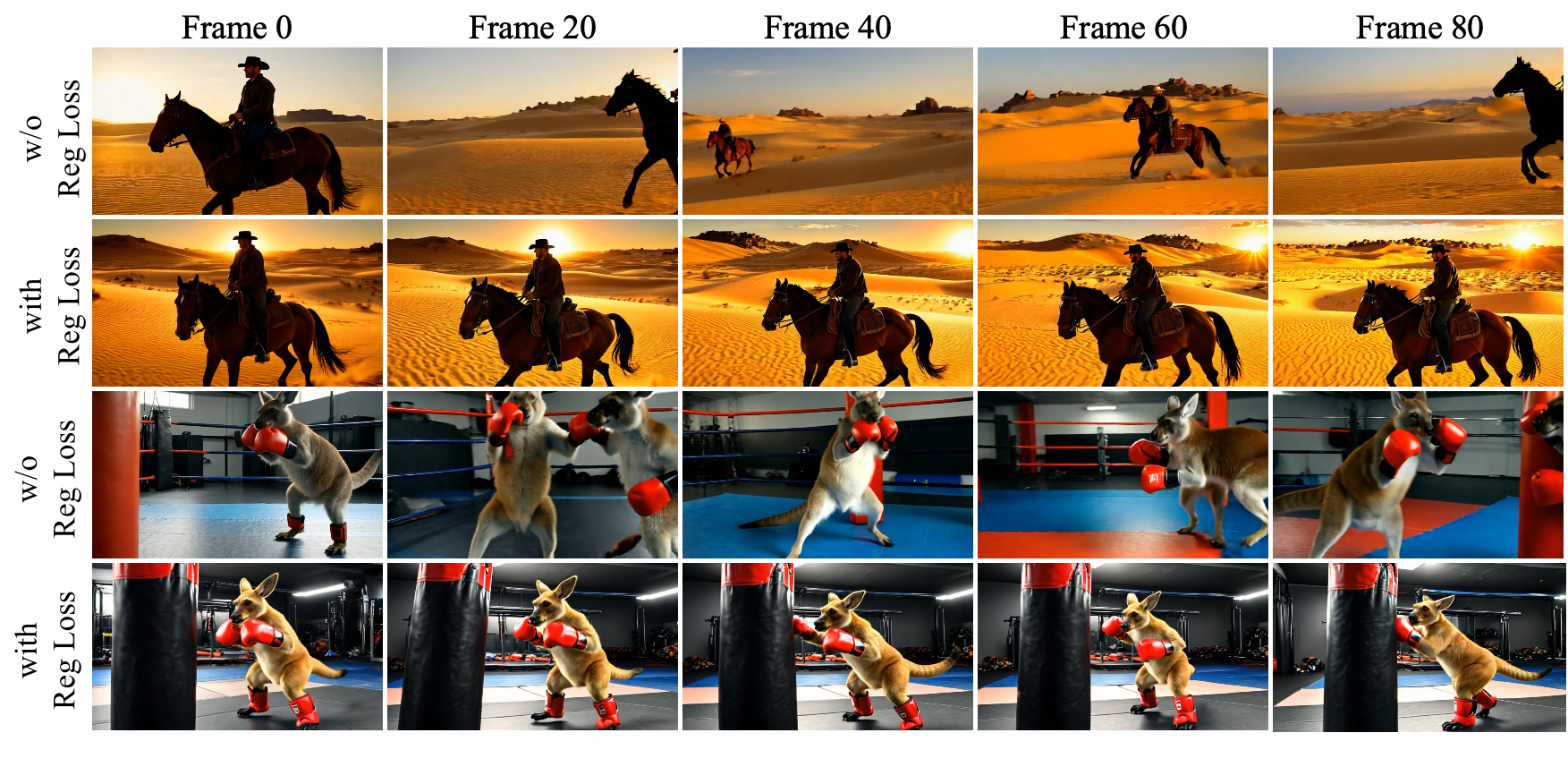 }
  \vspace{-2.5em}
  \caption{Visual results with and without the regularization loss. Removing the regularization loss can lead to persistent camera drift and character duplication. }
  \label{fig:reg_loss}
  \vspace{-1em}
\end{figure}

\noindent\textbf{Stability Enhancements.}
The relative temporal RoPE constraint and auxiliary regularization further strengthen robustness in extended streaming scenarios. Constraining temporal RoPE extrapolation mitigates positional distribution shift during long inference, yielding consistent quality gains. Meanwhile, trajectory-level regularization stabilizes motion dynamics and reduces long-horizon drift, as illustrated in \cref{fig:reg_loss}. Without this supervision, models exhibit gradual camera bias and occasional mode collapse.

\section{Conclusion}

We presented Hybrid Forcing, a unified architecture–optimization framework for efficient long-horizon streaming video generation. By integrating linear temporal attention for persistent global history aggregation with block-sparse sliding-window attention for efficient local refinement, our method systematically reallocates attention toward informative dependencies across time. Crucially, we further introduce a decoupled distillation strategy to fully unlock the potential of the hybrid attention mechanism, ensuring stable and effective long-horizon modeling. In addition, to mitigate temporal distribution shift and long-horizon mode collapse, we incorporate a consistent relative RoPE constraint and teacher-guided regularization for enhanced stability. Hybrid Forcing achieves state-of-the-art performance on both short- and long-form benchmarks and enables real-time, unbounded video generation at 29.5 FPS on a single H100 GPU.

\vspace{+1mm}
\noindent\textbf{Limitations}. Similar to prior streaming distillation approaches, our framework relies on the ODE model provided by Self-forcing~\cite{huang2025self}. Consequently, the student’s performance is inherently bounded by the provided model and the teacher’s generation quality. In addition, our throughput evaluation is conducted on high-performance GPUs, where the real-time performance may not be observed on low-end GPUs. We anticipate that integrating quantization and model compression techniques could narrow this gap, and we are actively exploring strategies to further reduce inference latency for broader deployment.

\clearpage  


%
%
\bibliographystyle{splncs04}
\bibliography{main}

\clearpage
\appendix                
In this supplementary file, we provide the following materials:

\begin{itemize}
    \item A complete description of our algorithm (see Sec. 3.5 in the main paper);
    \item The template used for enhancing VBench prompts (see Sec. 4.1 in the main paper);
    \item Detailed evaluation results on the short VBench benchmark and VBench-Long Benchmark (see Sec. 4.2 in the main paper);
    \item More qualitative comparisons under the 5-second and 30-second inference setting (see Sec. 4.2 in the main paper); (see Sec. 4.2 in the main paper);
    \item Cases of 10-minute videos (see Sec. 4.2 in the main paper).
\end{itemize}

We compile all visual comparison videos into a single MP4 file named \blueContent{comparison\_video.mp4}. In addition, we provide a complete video with all clips generated by our hybrid forcing model, saved as \blueContent{demo.mp4}. Both videos are included in the supplementary materials along with this PDF. Due to the size constraints of the supplementary materials, the videos are compressed at a lower bitrate.

\section{Algorithm details}

The details of training, inference, and hybrid attention are shown in \cref{alg:hybrid_forcing_train}, \cref{alg:hybrid_forcing_inference} and \cref{alg:hybrid_attention}, respectively.

\section{Template for VBench prompts enhancements}

Following previous works~\cite{huang2025self,yang2025longlive,ji2025memflow}, 
we enhance the text prompts in VBench using \textbf{Qwen/Qwen2.5-7B-Instruct}~\cite{bai2023qwen} 
together with the official prompt enhancement template. 
The prompt used to guide the enhancement is shown as follows:
\begin{tcolorbox}[breakable,colback=gray!5,colframe=gray!40,title=Prompt Enhancement Instruction]
\begin{lstlisting}[basicstyle=\ttfamily\footnotesize,breaklines=true]
You are a prompt engineer, aiming to rewrite user inputs into high-quality prompts for better video generation without affecting the original meaning.

Task requirements:
1. For overly concise user inputs, reasonably infer and add details...
2. Enhance the main features in user descriptions...
3. Output the entire prompt in English...
4. Prompts should match the user's intent...
5. Emphasize motion information...
6. Your output should have natural motion attributes...
7. The revised prompt should be around 80-100 words long.

Revised prompt examples:
1. Japanese-style fresh film photography, ...
2. Anime thick-coated illustration, ...
3. CG game concept digital art, ...
4. American TV series poster style, ...

I will now provide the prompt for you to rewrite...

\end{lstlisting}
\end{tcolorbox}

\begin{figure}[t!]
\begin{minipage}[t]{\textwidth}
  \begin{algorithm}[H]
    \caption{Hybrid Forcing Training}
    \small
    \begin{algorithmic}[1]
      \Require Denoise timesteps $\{\mathbf{t}\} = \{t_1, \dots, t_T\}$
      \Require Number of video chunks $N$
      \Require AR diffusion model $g_\theta$, Teacher diffusion model $g_{\phi}$
      \Require KV cache $\KVSet$ and Linear State $\LNSet$
      \Loop
        \State Initialize model output $\ModelOuput \gets []$
        \State Initialize noise input $\NoiseInput \gets []$
        \State Initialize KV cache $\KVSet \gets []$
        \State Initialize Linear State cache $\LNSet \gets []$
        \State Uniformly sample $s \sim len(\{\mathbf{t}\})$
        \For{$i = 1, \dots, N$}
          \State Initialize $\bm{x}^i_{t_T} \sim \mathcal{N}(0, I)$
          \State $\NoiseInput{\texttt{.append}}(\bm{x}^i_{t_T})$
          \For{$j = T, \dots, s$}
            \If{$t_j = t_s$}
              \State Enable gradient computation
              \State Set $\hat{\bm{x}}^i_{0} \gets g_\theta(\bm{x}^i_{t_{j}}; t_j, \KVSet, \LNSet)$ \Comment{Hybrid attention}
              \State $\ModelOuput{\texttt{.append}}(\hat{\bm{x}}^i_{0})$
              \State Disable gradient computation
              \State Cache $\KVOutput^i \gets g_\theta^\text{KV}(\hat{\bm{x}}^i_{0}; 0, \KVSet, \LNSet)$ \Comment{Hybrid attention}
              \If{$\KVSet$ is full}
                \State $\KVOutput^{evited} \gets \KVSet.\mathrm{pop}(0)$
                \State Update Linear State $\LNSet \gets \KVOutput^{evited}$ \Comment{Eq. (3) in the main paper}
              \EndIf
              \State $\KVSet{\texttt{.append}}(\KVOutput^i)$
            \Else
              \State Disable gradient computation
              \State Set $\hat{\bm{x}}^i_{0} \gets g_\theta(\bm{x}^i_{t_j}; t_j, \KVSet, \LNSet)$
              \State Sample $\bm{\epsilon} \sim \mathcal{N}(0, I)$
              \State Set $\bm{x}^i_{t_{j-1}} \gets \alpha_{t_{j-1}} \bm{\hat{x}}^i_{0} + \beta_{t_{j-1}} \bm{\epsilon}$ 
            \EndIf
          \EndFor
        \EndFor
        \If{$s = T$}
            \State $\mathbf{X}_{\phi} \gets \texttt{Rollout}(\NoiseInput,g_\phi,T)$
            \State $\mathcal{L}_{distill} = \mathcal{L}_{DMD} + \lambda \texttt{MSE}(\mathbf{X}_{\phi},\mathbf{X}_{\theta})$
        \Else
            \State $\mathcal{L}_{distill} = \mathcal{L}_{DMD}$
        \EndIf
        \State Update $\theta$ via $\mathcal{L}_{distill}$
      \EndLoop
    \end{algorithmic}
    \label{alg:hybrid_forcing_train}
  \end{algorithm}
\end{minipage}
\end{figure}

\begin{figure}[t!]
\begin{minipage}[t]{\textwidth}
  \begin{algorithm}[H]
    \caption{Autoregressive Diffusion Inference with Rolling KV Cache}
    \small
    \begin{algorithmic}[1]
      \Require Denoise timesteps $\{t_1, \dots, t_T\}$
      \Require Number of generated chunks $M$
      \Require AR diffusion model $g_\theta$
      \State Initialize model output $\ModelOuput \gets []$
      \State Initialize KV cache $\KVSet \gets []$
      \State Initialize Linear State cache $\LNSet \gets []$
      \For{$i = 1, \dots, M$}
        \State Initialize $\bm{x}^i_{t_T} \sim \mathcal{N}(0, I)$
        \For{$j = T, \dots, 1$}
          \State Set $\hat{\bm{x}}^i_{0} \gets g_\theta(\bm{x}^i_{t_j}; t_j, \KVSet, \LNSet)$
          \If{$j = 1$}
            \State $\ModelOuput{\texttt{.append}}(\hat{\bm{x}}^i_{0})$
            \State Cache $\KVOutput^i \gets g_\theta^\text{KV}(\hat{\bm{x}}^i_{0}; 0, \KVSet, \LNSet)$
            \If{$\KVSet$ is full}
              \State $\KVOutput^{evited} \gets \KVSet.\mathrm{pop}(0)$ \Comment{Cache eviction}
              \State Update Linear State $\LNSet \gets \KVOutput^{evited}$ \Comment{Eq. (3) in the main paper}
            \EndIf
            \State $\KVSet{\texttt{.append}}(\KVOutput^i)$
          \Else
            \State Sample $\bm{\epsilon} \sim \mathcal{N}(0, I)$
            \State Set $\bm{x}^i_{t_{j-1}} \gets \alpha_{t_{j-1}} \bm{\hat{x}}^i_{0} + \beta_{t_{j-1}} \bm{\epsilon}$
          \EndIf
        \EndFor
      \EndFor
      \State \Return $\ModelOuput$
    \end{algorithmic}
    \label{alg:hybrid_forcing_inference}
  \end{algorithm}
\end{minipage}
\end{figure}

\begin{figure}[t!]
\begin{minipage}[t]{\textwidth}
\begin{algorithm}[H]
    \small
    \caption{Hybrid Attention}
    \label{alg:hybrid_attention} 
    \begin{algorithmic}[1]
    \Require Matrices $Q, K, V \in \mathbb{R}^{N \times d}$.
    \Require Linear State Matrics $L\in \mathbb{R}^{d \times d}, H\in \mathbb{R}^{1 \times d}$.
    \Require Block sizes $b_q, b_{kv}$, hyper-parameters sparse Top $K$.
    \State \annotate{// Sparse Attention Part}
    \State Divide $Q$ to $T_m = N / b_q$ blocks $\{\vQ_i\}$ ;
    \State Divide $K, V$ to $T_n = N / b_{kv}$ blocks $\{\vK_i\},\{\vV_i\}$ ;
    \State {$P_c = {\rm Softmax}({\rm pool}(Q){\rm pool}(K)^\top / \sqrt{d})$ ; $ ~~~~~ $ Initialize $M_c=0$} ;
    \State {$M_c[i,j]=1$ if $P_c[i,j]\in{\tt TopK}(P_c[i,:], K)$} ;
    \For {$i=1$ {\bf to} $T_m$}
        \For {$j=1$ {\bf to} $T_n$} 
            \If {$M_c[i,j]=1$}
                \State {$\vS_{ij} = \vQ_i \vK_j^\top / \sqrt{d}$ ;
                \State $m_{ij} = {\rm max}(m_{i, j-1}, {\rm rowmax}(\vS_{ij}))$} ; 
                \State {$\vP_{ij}=\exp(\vS_{ij}-m_{ij})$} ;
                \State {$l_{ij}=e^{m_{i,j-1}-m_{ij}} l_{i,j-1} + {\rm rowsum}(\vP_{ij})$ ; 
                \State $\vO_{ij}^s = {\rm diag}(e^{m_{i,j-1}-m_{ij}}) \vO_{i,j-1}^s + \vP_{ij} \vV_j$} ;
            \EndIf
        \EndFor
        \State $\vO_i^s={\rm diag}(l_i^{T_n})^{-1}\vO_{i,T_n}^s$ ;
        \State $O^{local} = \{\vO^s_i\}$
    \EndFor
    \State \annotate{// History Linear Attention Part}
    \State $O^{\text{history}} =
    \text{Linear}\!\left(
    \frac{\text{RoPE}(\phi(Q)) L}
         {\phi(Q) H}
    \right)$
    \State \textbf{return} $O=O^{local} + O^{hitory}$ ;
    \end{algorithmic}
\end{algorithm}
\end{minipage}
\end{figure}

\section{Detailed evaluation results on the short VBench benchmark and VBench-Long Benchmark}

\begin{figure}[tb]
  \centering
  \includegraphics[width=\linewidth]{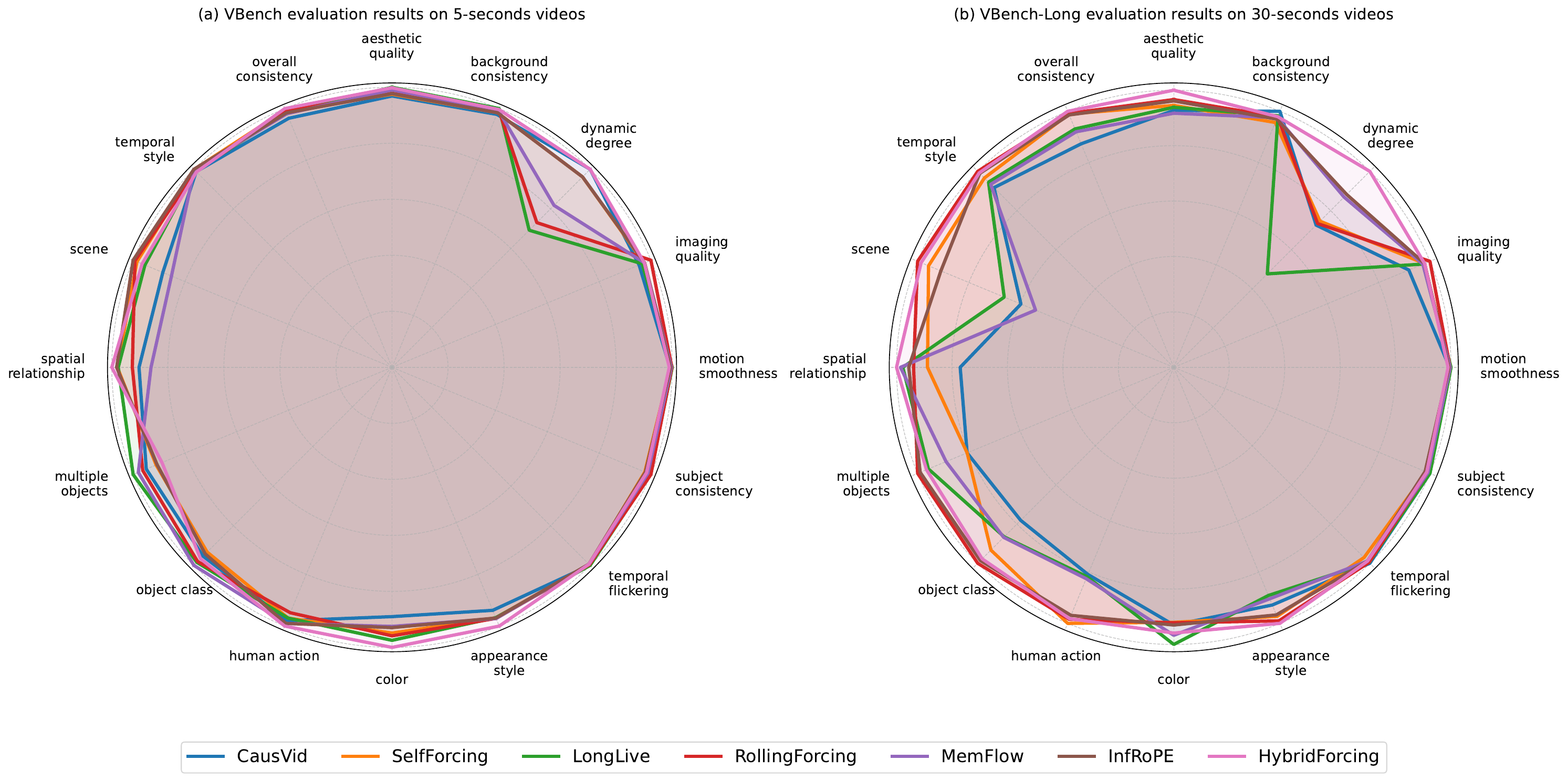 }
  \caption{(a) VBench evaluation results over all 16 metrics on 5-seconds videos. (b) VBench-long evaluation results over all 16 metrics on 30-seconds videos. Values are normalized by the max value for better visualization effect.}
  \label{fig:vbench_long}
\end{figure}

In \cref{fig:vbench_long}, we present the detailed evaluation results on VBench and VBench-Long for representative few-step autoregressive models across all 16 VBench metrics. Overall, our hybrid forcing method consistently outperforms its competitors, with particularly notable improvements in semantic-related metrics. 

Specifically, under the 5-second evaluation setting, our hybrid forcing achieves leading performance on several metrics, including dynamic degree, appearance style, color, spatial relationship, and overall consistency, while maintaining comparable results on the remaining metrics. 
Under the more challenging 30-second generation setting, hybrid forcing demonstrates even clearer advantages over existing autoregressive methods. Our method ranks first or second on most metrics, with especially pronounced improvements in aesthetic quality, dynamic degree, and spatial relationship. These results suggest that hybrid forcing better preserves semantic coherence and visual consistency when generating longer videos.

\begin{figure}[tb]
  \centering
  \includegraphics[width=\linewidth]{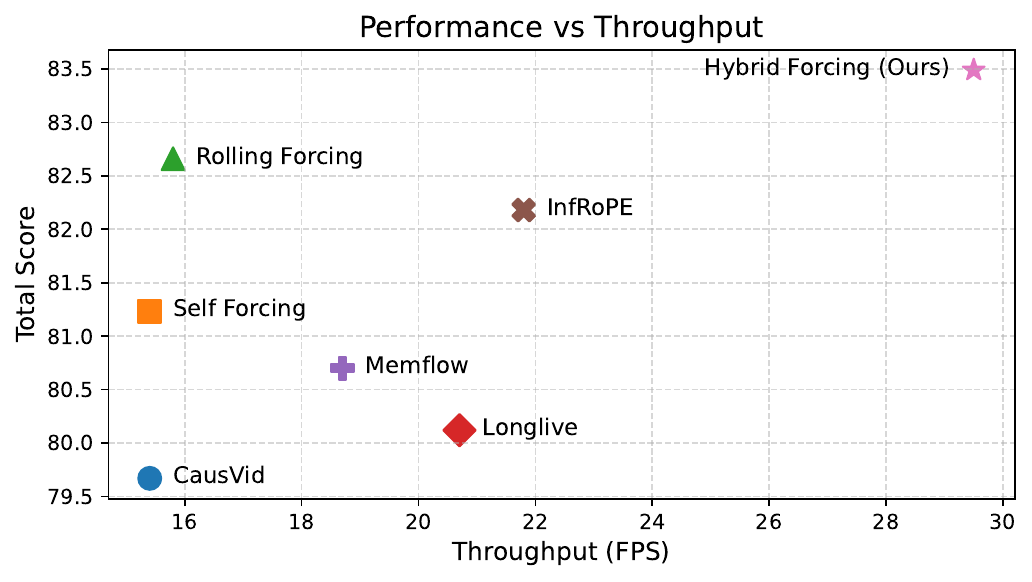 }
  \caption{Performance (total score vs. throughput) comparison across seven methods on 30-seconds video inference. Our method achieves superior quality and efficiency.}
  \label{fig:vbench_performance}
\end{figure}

Furthermore, as illustrated in \cref{fig:vbench_performance}, our method not only achieves the best overall \textit{Total Score} under the 30-second setting, but also attains the highest throughput among all compared methods, highlighting its favorable balance between generation quality and efficiency.

\section{More qualitative comparisons under the 5-second and 30-second inference setting}

\begin{figure}[tb]
  \centering
  \includegraphics[width=\linewidth]{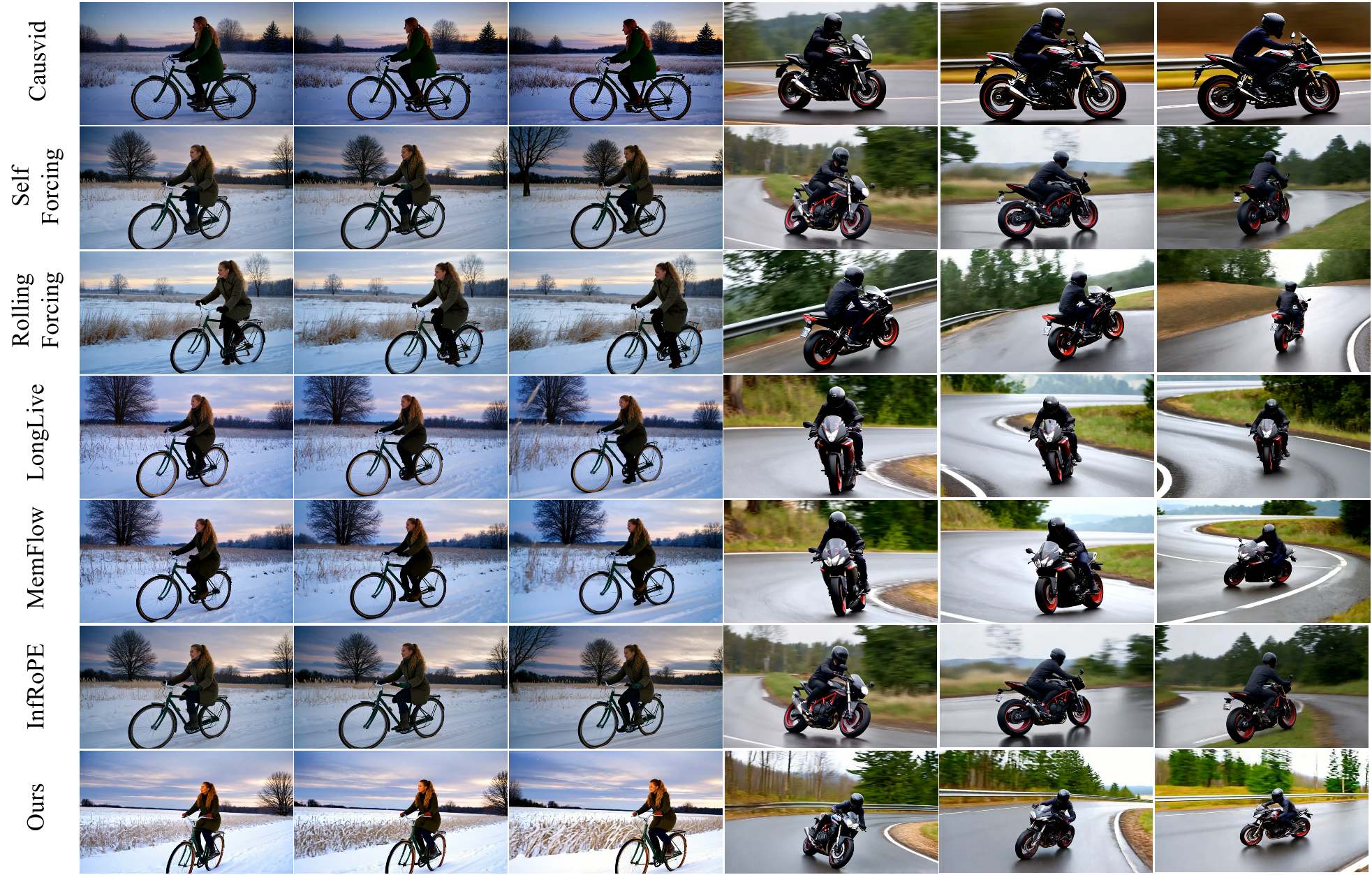 }
  \caption{Qualitative comparison on 5-seconds video.}
  \label{fig:v_short1}
\end{figure}

\begin{figure}[tb]
  \centering
  \includegraphics[width=\linewidth]{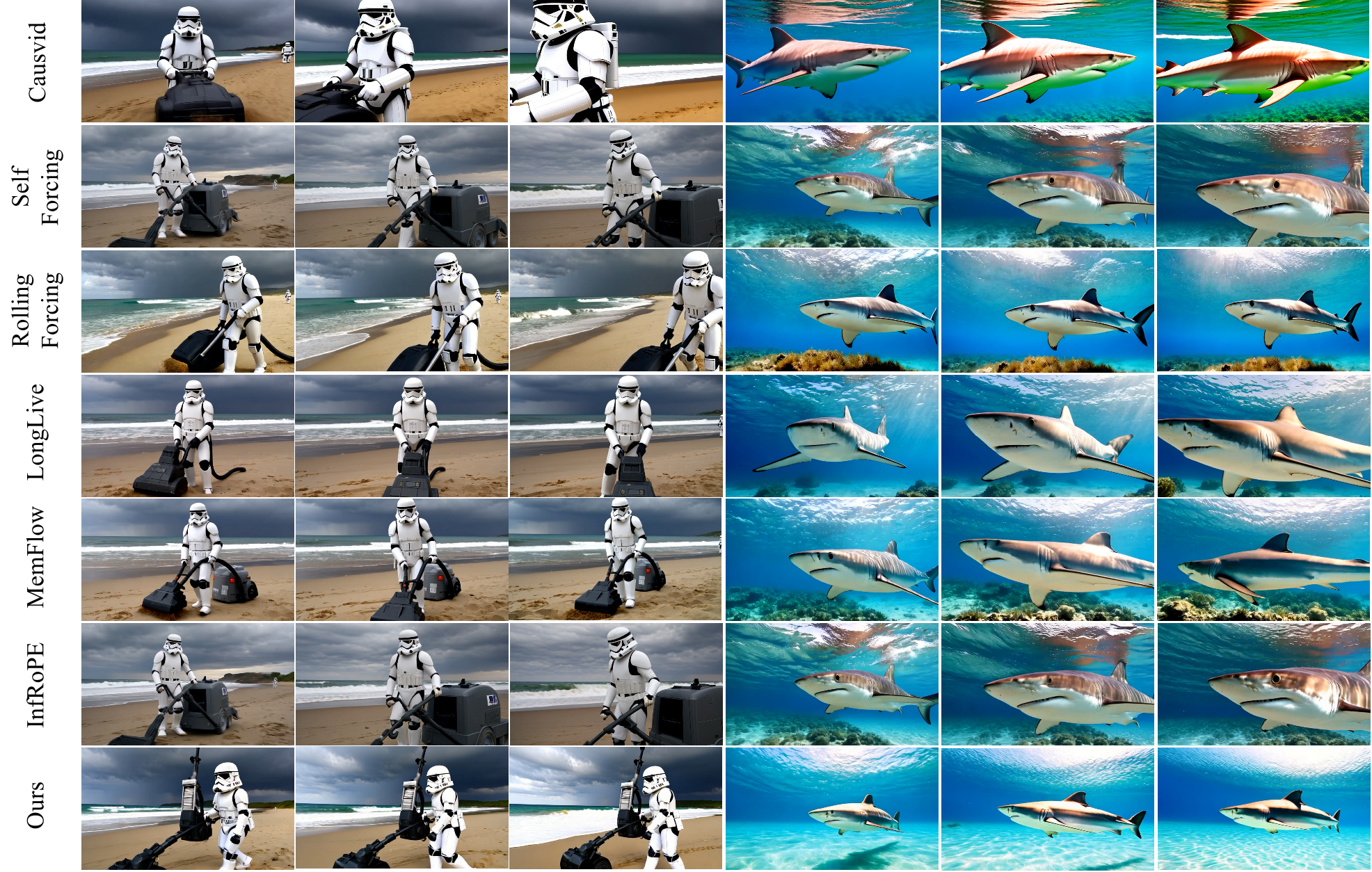 }
  \caption{Qualitative comparison on 5-seconds video.}
  \label{fig:v_short2}
\end{figure}

In \cref{fig:v_short1} and \cref{fig:v_short2}, we present additional qualitative comparisons for 5-second video generation. As shown in the figures, \textit{LongLive} and \textit{MemFlow} exhibit limited motion dynamics, particularly in the bike-riding and motorcycle-riding cases. 
This issue becomes more apparent when inspecting the MP4 file in the \blueContent{00:02–00:32} segment of \blueContent{comparison\_video.mp4} in the supplementary materials., where the bicycle and the motorcycle nearly remain stationary throughout the sequence. In addition, the motion ranges of objects such as the stormtrooper and the shark evolve more slowly compared to other methods, indicating weaker temporal dynamics. In contrast, our hybrid forcing approach produces more natural motion patterns while maintaining stronger visual fidelity and temporal consistency. Please refer to the corresponding \blueContent{comparision\_video.mp4} in our supplementary materials for full video comparisons.

\begin{figure}[tb]
  \centering
  \includegraphics[width=\linewidth]{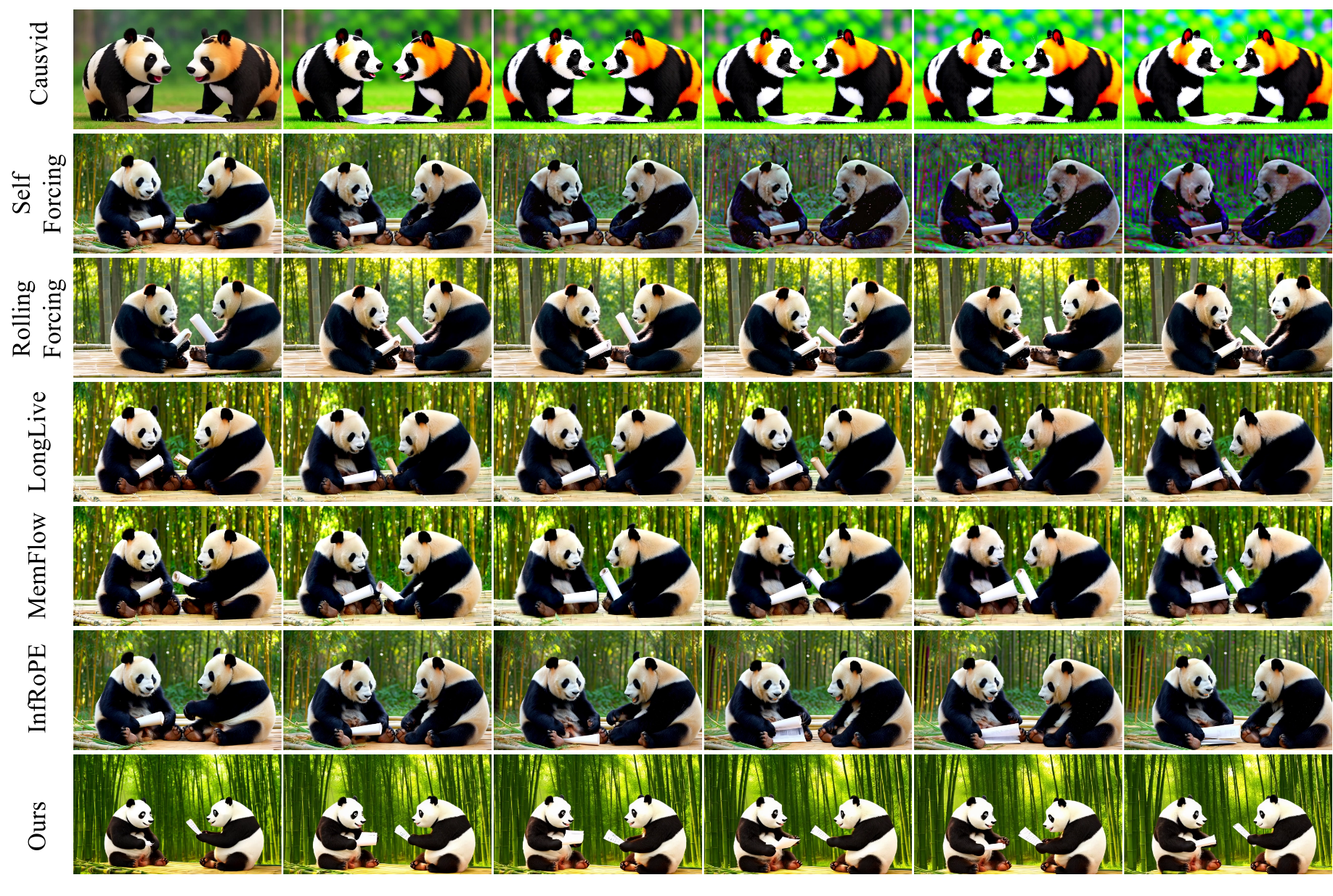 }
  \caption{Qualitative comparison on 30-seconds video.}
  \label{fig:v_long1}
\end{figure}

\begin{figure}[tb]
  \centering
  \includegraphics[width=\linewidth]{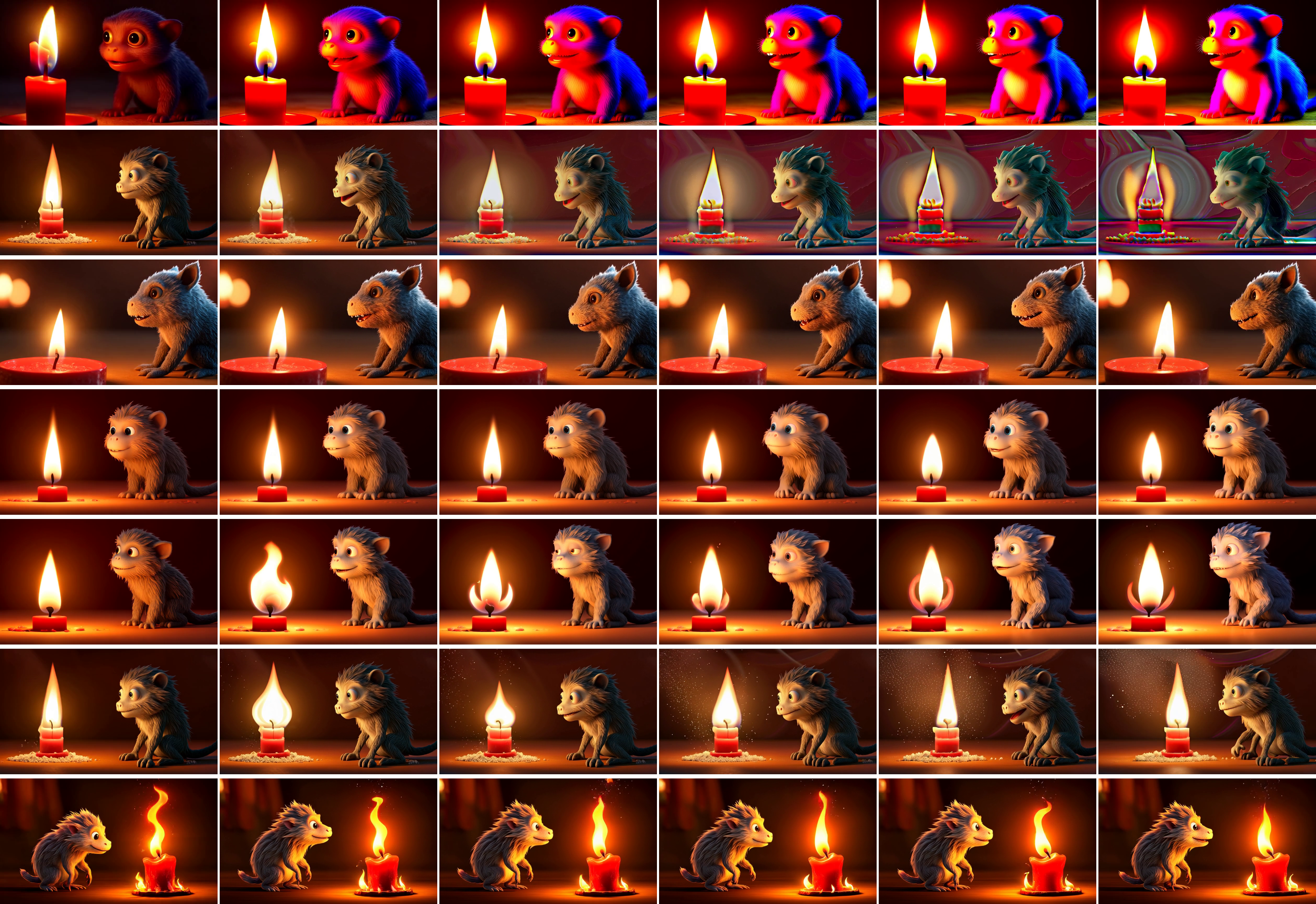 }
  \caption{Qualitative comparison on 30-seconds video.}
  \label{fig:v_long2}
\end{figure}

\begin{figure}[tb]
  \centering
  \includegraphics[width=\linewidth]{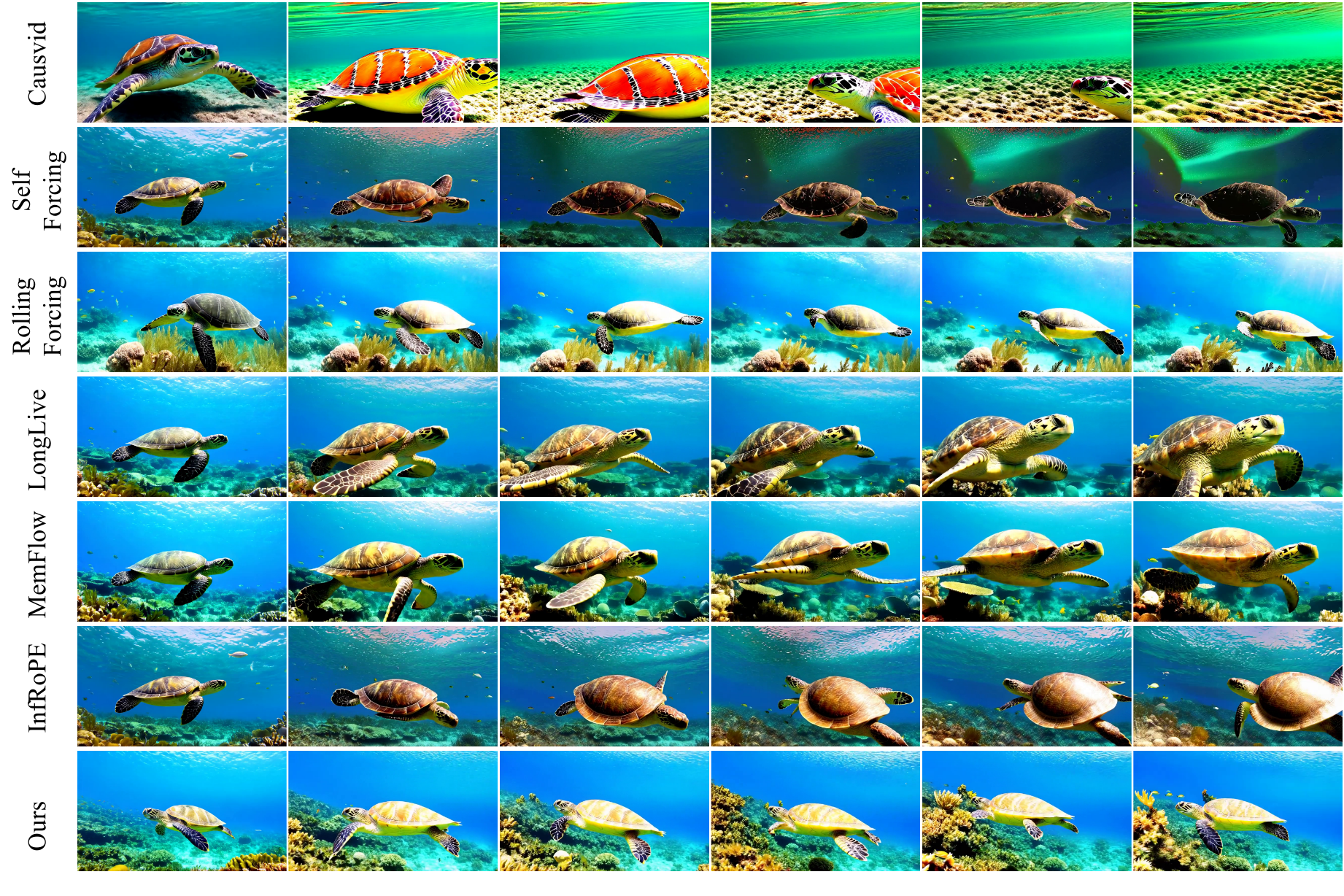 }
  \caption{Qualitative comparison on 30-seconds video.}
  \label{fig:v_long3}
\end{figure}

In \cref{fig:v_long1}, \cref{fig:v_long2}, and \cref{fig:v_long3}, we provide qualitative comparisons for 30-second video generation. As illustrated in the figures, \textit{CausVid}, \textit{Self-Forcing}, and \textit{InfRoPE} struggle to maintain stable generation over long durations, with noticeable visual degradation emerging after approximately 10 seconds. In extreme cases, severe structural artifacts appear; for example, in the turtle video, the turtle’s tail gradually deforms and visually merges with the head in the middle of the sequence. While other methods demonstrate relatively stable performance, our hybrid forcing method achieves comparable or better visual quality while achieving the highest generation throughput. Specifically, our method achieves $\times1.42$, $\times1.57$, and $\times1.86$ higher throughput than LongLive, MemFlow, and Rolling Forcing, respectively, demonstrating a favorable balance between long-horizon generation quality and computational efficiency. Please refer to the \blueContent{00:34–02:22} segment of  \blueContent{comparison\_video.mp4} in the supplementary materials for full video comparisons.

\section{Cases of 10-minute videos}

\begin{figure}[tb]
  \centering
  \includegraphics[width=\linewidth]{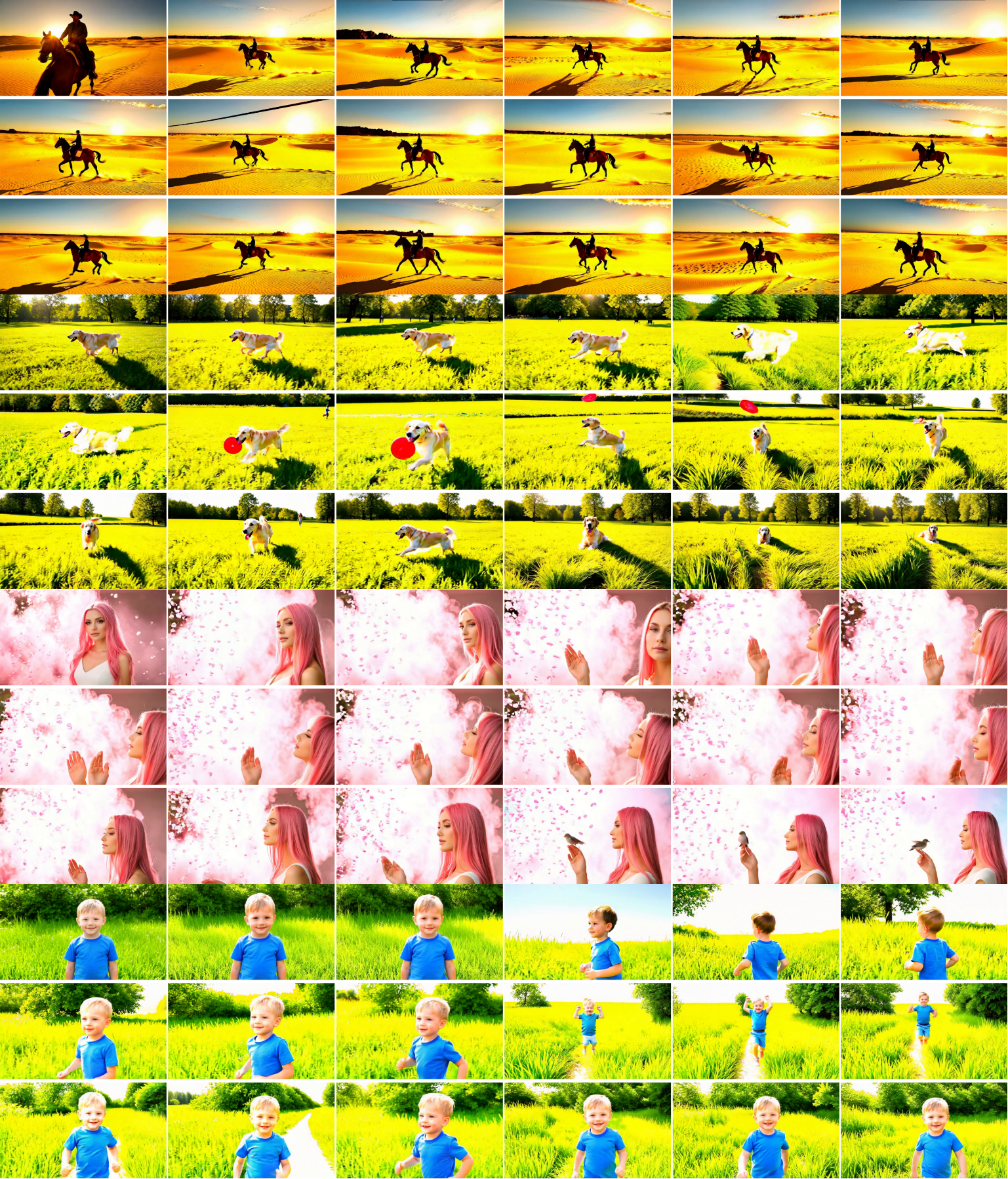 }
  \caption{Visual result of 10 minutes video.}
  \label{fig:10min_case1}
\end{figure}

We present examples of single-prompt 10-minute video generation in \cref{fig:10min_case1}. As illustrated in the figure, both character identity and background appearance remain largely consistent throughout the entire sequence, even after 10 minutes of continuous generation. This demonstrates the strong long-term temporal stability of our approach. Please refer to the \blueContent{02:25–02:42} segment of \blueContent{comparison\_video.mp4} in the supplementary materials for videos with $32\times$ speed-up.

Combined with our high generation throughput of \textbf{29.5 FPS}, our proposed method is well suited for latency-sensitive application scenarios, including real-time livestreaming, virtual avatar systems, and interactive world models, where both long-duration consistency and high inference speed are essential.

\end{document}